\documentclass[11pt]{article}
\usepackage[utf8]{inputenc}
\usepackage{amsfonts}
\usepackage{amsmath}
\usepackage{amsthm}
\usepackage{color}
\usepackage{newtxtext}
\usepackage{subcaption}
\usepackage{graphicx}
\usepackage{algorithm}
\usepackage{algorithmic}
\usepackage{etoolbox}  % patch def of algorithmic environment
\makeatletter
\patchcmd{\algorithmic}{\addtolength{\ALC@tlm}{\leftmargin} }{\addtolength{\ALC@tlm}{\leftmargin}}{}{}
\makeatother
\usepackage{array}
\usepackage{eqparbox}

\usepackage[margin=1in]{geometry}
\usepackage{mathtools}
\usepackage{booktabs}
\usepackage{authblk}
\usepackage{url}
\graphicspath{{figs/}}

\usepackage{graphicx}
\usepackage{booktabs} % for professional tables
\usepackage{subcaption}
\usepackage{multirow}
\usepackage{csquotes}
\usepackage{listings}
\usepackage{array}
\usepackage{tabularx}
\usepackage{multirow}
\usepackage{multicol}
\usepackage{hhline}
\usepackage{caption}
\usepackage{comment}
\usepackage{float}
\usepackage{lineno}
\usepackage{hyperref}
\title{Dynaformer: A Deep Learning Model for Ageing-aware Battery Discharge Prediction}
\author[1,3]{Luca Biggio}
\author[3]{Tommaso Bendinelli}
\author[4]{Chetan Kulkarni}
\author[2]{Olga Fink}
\affil[1]{Data Analytics Lab, ETH, Zürich, Switzerland}
\affil[2]{Laboratory for Intelligent Maintenance and Operations Systems, EPFL, Switzerland}
\affil[3]{CSEM SA, Alpnach, Switzerland}
\affil[4]{KBR, Inc., NASA Ames Research Center, Mountain View, CA 94035, USA}
\date{}

\begin{document}
\maketitle

\begin{abstract}
Electrochemical batteries are ubiquitous devices in our society.
When they are employed in mission-critical applications, the ability to precisely predict the end of discharge under highly variable environmental and operating conditions is of paramount importance in order to support operational decision-making. 
While there are accurate predictive models of the processes underlying the charge and discharge phases of batteries, the modelling of ageing and its effect on performance remains poorly understood. Such a lack of understanding often leads to inaccurate models or the need for time-consuming calibration procedures whenever the battery ages or its conditions change significantly. This represents a major obstacle to the real-world deployment of efficient and robust battery management systems.
In this paper, we propose for the first time an approach that can predict the voltage discharge curve for batteries of any degradation level without the need for calibration. In particular, we introduce Dynaformer, a novel Transformer-based deep learning architecture which 
is able to \emph{simultaneously} infer the ageing state from a limited number of voltage/current samples and predict the full voltage discharge curve for real batteries with high precision.
Our experiments show that the trained model is effective for input current profiles of different complexities and is robust to a wide range of degradation levels. In addition to evaluating the performance of the proposed framework on simulated data, we demonstrate that a minimal amount of fine-tuning allows the model to bridge the simulation-to-real gap between simulations and real data collected from a set of batteries. The proposed methodology allows for accurate planning and control over missions characterized by complex usage profiles. Moreover, it enables the utilization of battery-powered systems until the end of discharge in a controlled and predictable way, thereby significantly prolonging the operating cycles and reducing costs.

\end{abstract}

\noindent
\section*{Introduction}
Since their introduction to the market in 1991 \cite{horiba2014lithium}, lithium-ion batteries have had a significant impact on modern society. They have been at the heart of revolutions in several technological areas, including portable electronics, autonomous systems, wireless communications, and electric vehicles. Furthermore, they are still considered key strategic components for replacing fossil fuels with greener energy sources \cite{opp}.
At a time when the quest for carbon neutrality is pressing and the electric conversion of land and air vehicles is progressing rapidly, the need for efficient battery management tools to reliably model the behaviour of batteries under varying operating and environmental conditions has never been so urgent \cite{opp3}.\\
Today, an open challenge of great importance in the context of battery management is the problem of accurate End-of-Discharge (EoD) prediction, i.e. inference and monitoring of the time left until a battery reaches its discharge point. The implications of a reliable and precise EoD estimation method can be significant. In the context of electric vehicles, for example, such a method could alleviate the problem of range anxiety \cite{opp5}, which has been one of the major obstacles to a wider acceptance and integration of electric vehicles into the market \cite{opp4}. While capturing the dynamics of the discharge is already a challenging task, the problem of EoD prediction is further exacerbated by the fact that batteries degrade with time. Battery ageing represents a deviation from nominal operating conditions, which in turn results in decreased performance due to a loss of capacity and increased impedance. A precise physical model of the ageing process is difficult to formulate due to its inherent complexity and the multiple unpredictable sources contributing to its development \cite{opp2}. Yet inferring the level of ageing reached by the battery at a certain point in its lifespan is crucial, since its discharge point can be highly influenced by the current level of degradation.\\
The methods to tackle EoD predicton and ageing inference investigated in the literature fall mainly into two categories: model-based \cite{phy1,phy2,phy3,phy4,phy5,phy6} and data-driven \cite{ml1,ml2,ml4,ml5,ml6}. The first class of methods is either based on the characterization of the physical mechanisms at the core of the battery's internal processes \cite{phy6} or on the construction of an ideal model of the battery which reflects its macroscopic behaviour, for instance in the form of an equivalent circuit \cite{phy2}. Both approaches require extensive domain knowledge and have their own limitations. Despite their good performance, physics-based models are typically computationally expensive due to the complex differential equations involved in their formulation and are thus not well suited for online applications \cite{phy6}. Furthermore, a precise physical characterization of intrinsically complex phenomena such as ageing and degradation necessitates a very granular modeling of the battery's underlying processes, which is often not feasible in practice.  Approaches based on equivalent circuits, on the other hand, are often much faster, albeit at the cost of providing less accurate predictions. An important property of model-based methods is that they are generally highly interpretable by human experts since their design stems from first-principle physical considerations. Nonetheless, biases, missing physics, or erroneous assumptions in the model formulation can lead to significant discrepancies with real observations. Thus, model-based approaches are not easily applicable when the system of interest is subject to heterogeneous and variable operating conditions.
\\  Research on the second class of methods,  the data-driven approaches, has been progressing steadily over the last few years, motivated by the potential of machine learning techniques to overcome some of the limitations of model-based approaches. Machine learning models -- including neural networks, Gaussian processes, random forests, and support-vector machines -- have been successfully applied to the problems of EoD prediction and degradation inference (see \cite{opp4} for an extensive review). Given enough data, once trained, such models are able to make accurate predictions in negligible time with little to no prior knowledge on the battery's underlying discharge and degradation processes. These features make them particularly well suited to modeling complex nonlinear phenomena that are difficult to describe mathematically or that result in computationally expensive approaches, ageing being a particularly relevant example. Nevertheless, the application of data-driven techniques is hindered by a number of limiting factors. First, state-of-the-art models require large labeled training datasets and collecting such large datasets from electrochemical batteries has been notoriously difficult and very time-consuming \cite{opp4}. Second, the EoD prediction task typically requires processing of very long time series and involves predictions over long time horizons. This is a feature that makes most data-driven models ineffective and computationally inefficient. Third, modern data-driven techniques, such as neural networks, are often considered black-box methods and their output, contrary to model-based techniques, is typically very difficult to interpret by domain experts.\\
First steps towards the solutions to some of the aforementioned drawbacks have been taken in \cite{ml6} and \cite{ml3}, from which our work draws inspiration. In \cite{ml6}, the authors explore the application of transfer learning techniques to mitigate the need for large datasets to train their deep convolutional architecture on the task of battery state-of-health estimation. However, the authors mainly focus on the problem of capacity fade estimation and do not consider EoD prediction in their experiments. In \cite{ml3}, a hybrid model incorporating physics-based modeling of lithium-ion batteries and data-driven components is developed for EoD forecasting. The authors take ageing into account by performing parameter inference of two degradation parameters and, after this fitting step is terminated, they proceed with EoD prediction. Thanks to its hybrid nature, the proposed approach inherits the high interpretability of the physics-based model it is based on. However, although the authors explicitly account for ageing in their hybrid model, their method needs to be partially re-calibrated whenever the ageing condition of the battery deviates significantly.\\ %Additionally, the resulting approach cannot be easily adapted to real data since it is rigidly constrained by the chosen physics-based model. \\
Despite the increasing interest in battery management and the high-quality solutions resulting from the previously described research efforts, models that can perform EoD prediction and ageing inference \emph{simultaneously} and that can be flexibly applied to real-world, battery-operated devices loaded with heterogeneous current profiles are still missing and very much needed. 
\paragraph{Contributions.}
In this work, we propose Dynaformer, a novel machine learning algorithm for ageing-aware EoD prediction with the following main properties: 1) based only on a small window of current/voltage observations, it can accurately predict the full voltage discharge curve and the degradation state simultaneously; 2) it can efficiently process very long time series; 3) it is able to process both constant and variable load profiles of an arbitrary level of complexity; 4) it can be efficiently adapted to real battery data by combining a pre-training phase on simulated data and a transfer learning procedure; 5) its feature space is highly interpretable and encodes meaningful information on the ageing level (contrary to most data-driven approaches).\\
Dynaformer is based on an encoder-decoder neural network architecture whereby the encoder learns to extract physics-based information from observations and the decoder performs EoD prediction. In particular, ageing inference is implicitly performed by the encoder given only the initial part of the current/voltage profiles. Conditioned on the encoder output, the decoder predicts the full voltage discharge trajectory resulting from an arbitrary current load profile given as input. We empirically demonstrate that the first two principal components of the latent representation extracted by the encoder are highly correlated with two degradation parameters, namely the total amount of available lithium ions $q_{max}$ and the internal resistance $R_0$, characterizing the ageing state of the battery. This feature confers a high level of interpretability to our approach and, contrary to previous works, enables it to perform ageing inference and EoD prediction simultaneously and in an end-to-end fashion.\\
To cope with the problem of long-range prediction, Dynaformer leverages a novel architecture based on Transformers \cite{tr}, whose design draws inspiration from recently developed solutions in computer vision \cite{vit}. The resulting model enables the processing of very long time series, keeping the computation time and memory footprint sufficiently small for real applications. \\
Due to the lack of sufficiently large representative training datasets, we employ a high-fidelity, physics-based simulator \cite{2021_nasa_prog_models} to generate -- offline -- a very large dataset covering a wide range of degradation conditions and ageing stages of a Li-ion battery.
%\begin{figure}[H]
%    \centering
%    \includegraphics[scale = 0.4]{varyingQ (5).pdf}
%    \includegraphics[scale = 0.4]{varyingR (3).pdf}
%    \caption{\textbf{Effect of varying the degradation parameters on the voltage discharge curve}. (Left) Varying $q_{max}$ and keeping $R_0$ fixed; (Right) Varying $R_0$ and keeping $q_{max}$ fixed. All the curves in the plots result from the same 1A constant current profile.}
%    \label{fig:degradation_params}
%\end{figure}
After pre-training on such data, we empirically show that our model is flexible enough that simply fine-tuning it on a small subset of real data allows us to bridge the \emph{simulation-to-real gap}, representing the missing physics information that the simulator has not explicitly been designed to capture. \\
%To address the problem of the lack of representative training data, we employ a high-fidelity physics-based simulator \cite{2021_nasa_prog_models} to generate -- offline -- a very large dataset that we use to train our model. The generated dataset is carefully designed to cover a wide range of degradation conditions, representing several ageing stages of a Li-ion battery. Such a large-scale training phase results in a model that is robust to a large set of degradation states that can be encountered in real data. \\
%Despite being trained on simulated data, we empirically show that our model is flexible enough that simply fine-tuning it on a small subset of real data, allows us to fill the \emph{simulation-to-real-gap} representing the missing physics information that the simulator has not explicitly been designed to capture.  \\
Overall, Dynaformer shows very good generalization performance to unseen degradation conditions and it is able to handle arbitrarily complex load profiles, even when these are chosen significantly outside the training distribution. Our model represents a powerful new tool for battery management, enabling enhanced ageing-aware control over the EoD of battery-powered systems subject to highly heterogeneous operating conditions. More generally, the proposed method is applicable to any long-term prediction task in which the estimation of the variable of interest depends on a preliminary inference stage. This includes, for example, the prediction of the remaining useful lifetime \cite{10.3389/frai.2020.578613} of a mechanical component, but also, more broadly, the estimation of time-dependent variables in dynamical system with unknown parameters.

%%The proposed method is general and its application can be envisioned for all those frameworks in which a certain task of interest, such as for example the prediction of the remaining useful lifetime of a mechanical component, depends on a preliminary inference stage of the underlying degradation dynamics. 

\section*{Methods}\label{methodology}
\begin{figure}[H]
    \centering
    \includegraphics[scale = 0.38]{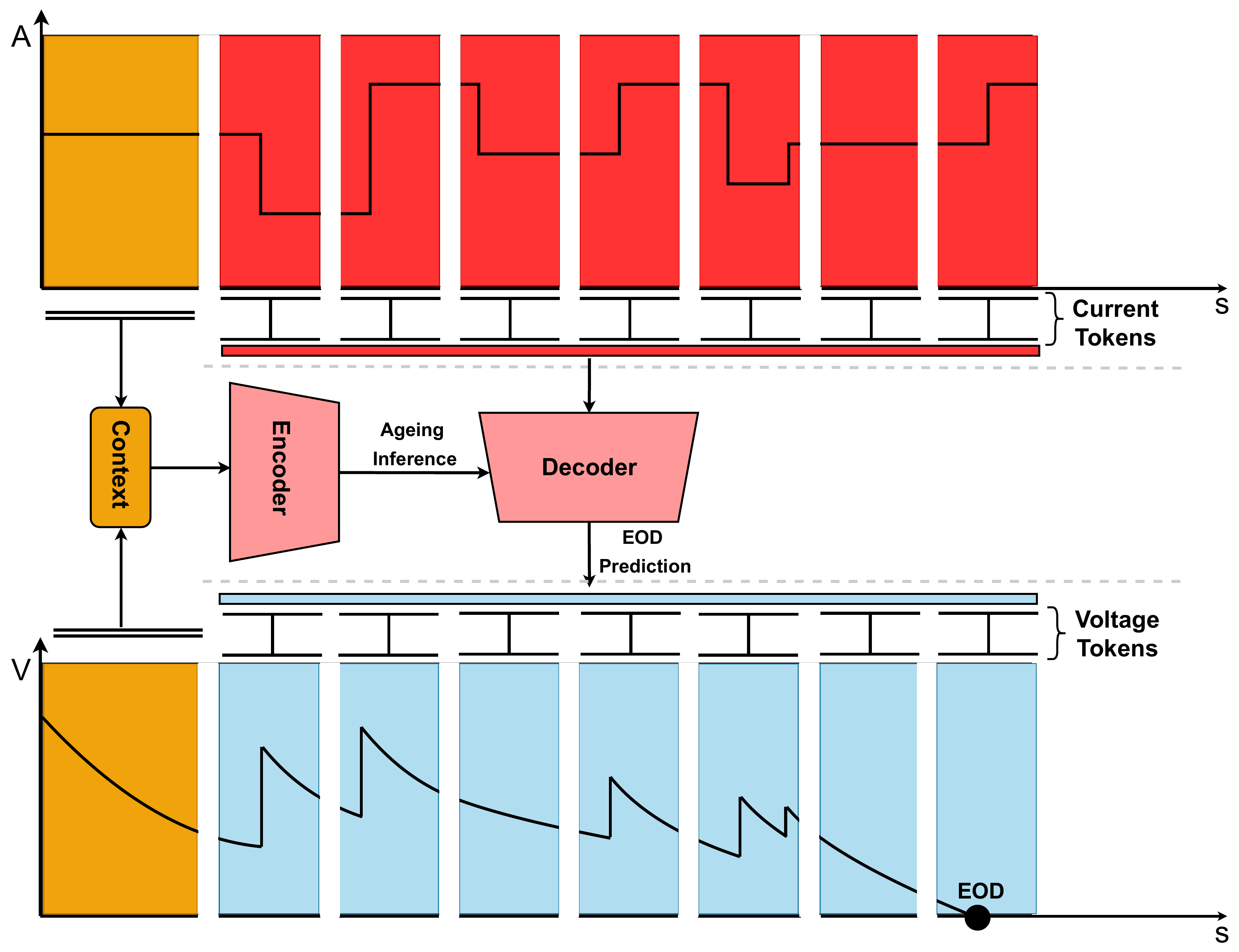}    \caption{\textbf{Representation of the components of Dynaformer.} %Arrows show the information flow among the various components described in this paper. 
    First, a context embedding vector is extracted from the initial parts of the current profile and the discharge curve. The context is then processed by an encoder, which extracts information about the level of degradation and ageing. The Dynaformer decoder receives two different inputs: first, the input current profile and, second, the output of the encoder that acts as a conditioning vector. The input current profile is usually represented by a long time series, making it hard for standard approaches to effectively process it. To cope with this aspect, the sequence is split into multiple sub-sequences, each of which is treated as a single token by the decoder. The output of the decoder is a sequence of voltage tokens that, once concatenated, represent the complete voltage trajectory.}
    \label{fig:mainfigure}
\end{figure}

The goal of this work is to develop a data-driven model that -- given a new, arbitrarily complex current profile and a small number of voltage/current observations at the beginning of the discharge trajectory -- is able to predict the remaining part of the voltage curve until the EoD (see Fig. \ref{fig:mainfigure}). This task is made substantially more complicated by the ageing effect, which significantly impacts the shape of the discharge trajectories. Therefore, in addition to the EoD prediction, the model also needs to perform ageing inference in order to determine the degradation level of the battery at that specific cycle. In this work, we propose to address EoD prediction and ageing inference \emph{simultaneously} and in an end-to-end manner. This is achieved by generating a large and diverse set of simulated training data and by designing a modular machine learning algorithm capable of performing ageing-aware EoD prediction and of efficiently handling very long time series. To bridge the simulation-to-real gap between simulated and real battery data, we adopt a transfer-learning procedure that allows our model to be successfully applied to data from a set of actual battery models. In the following paragraphs, we introduce the main building blocks of our approach, namely the data generation as well as the machine learning method and its fine-tuning to close the simulation-to-real gap.

%Fig \ref{fig:mainfigure} provides a schematic description of the main components of Dynaformer. \\

\noindent

%% how can this be achieved: fusion between model-based and data driven
%but still retaining flexibility -- not rigid

\subsection*{Data Generation}
\begin{figure}[H]
\centering
\begin{subfigure}{1\textwidth}
\centering
    \includegraphics[scale = 0.4]{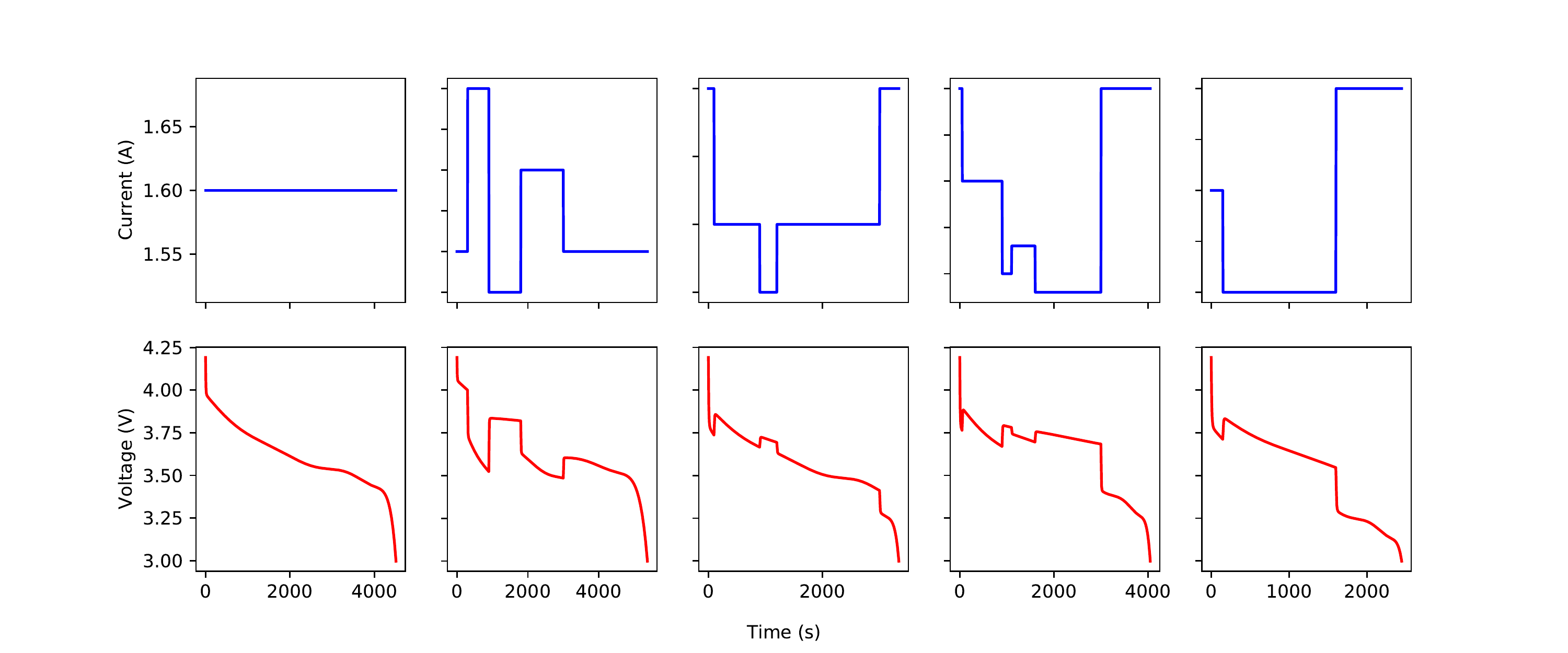}
    \caption{Shown are five examples of current load profile (Top) and the corresponding voltage discharge curves (Bottom) with degradation parameters ($q_{max}$, $R_0$) fixed.}
    \label{fig:constant_variable}
\end{subfigure}\\
\begin{subfigure}{1\textwidth}
\centering
    \includegraphics[scale = 0.35]{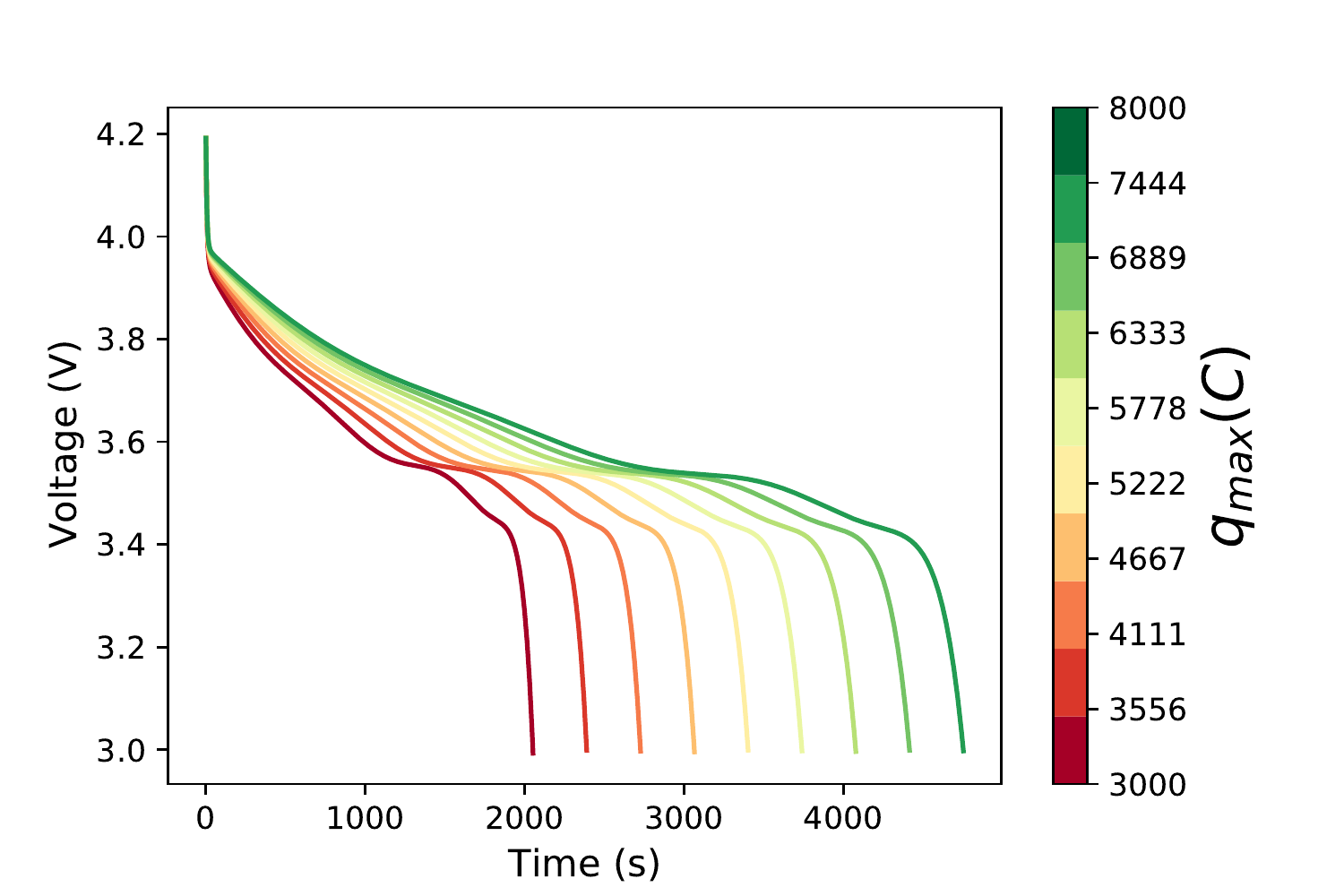}
    \includegraphics[scale = 0.35]{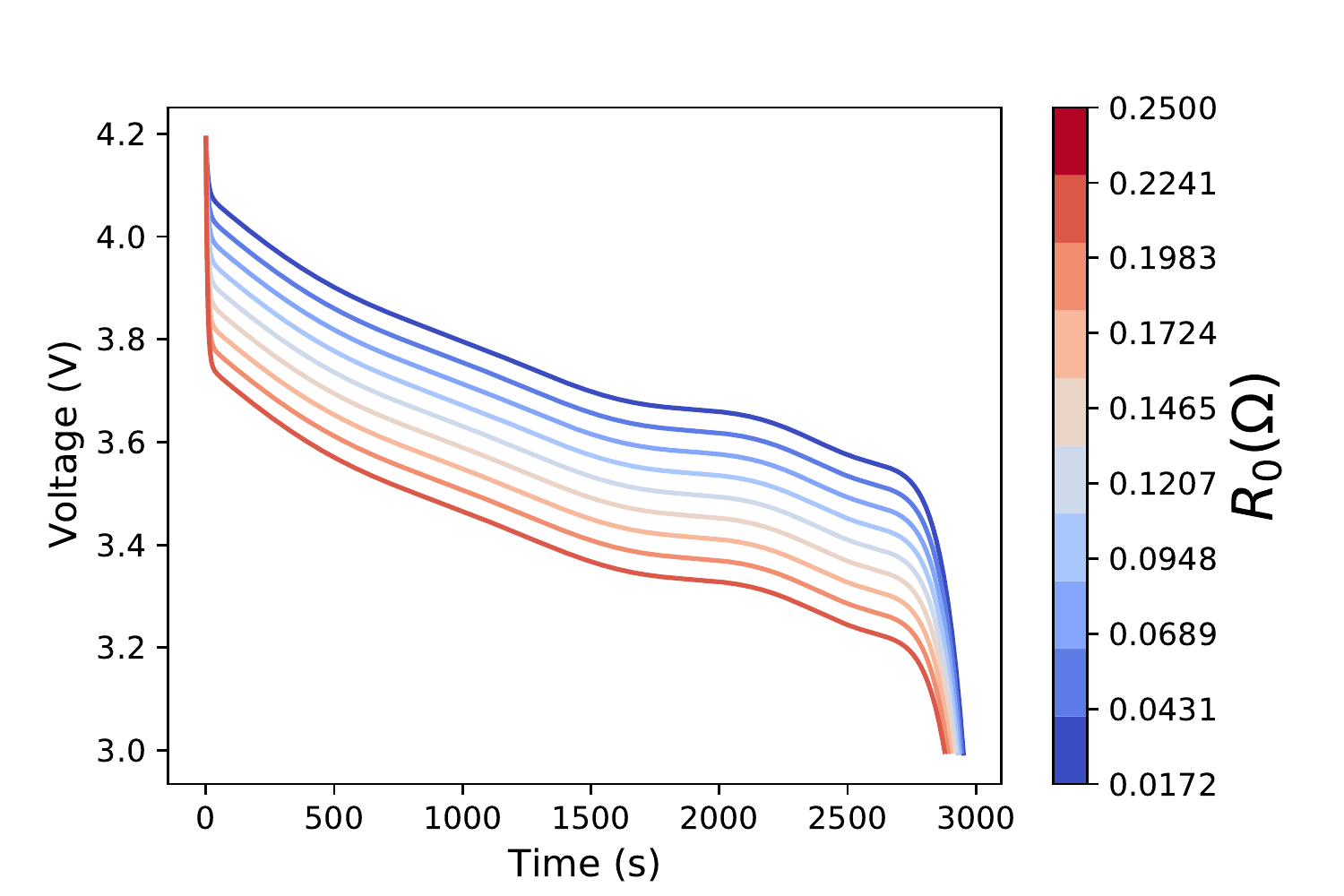}
    \caption{(Left) Varying $q_{max}$ and keeping $R_0$ fixed; (Right) Varying $R_0$ and keeping $q_{max}$ fixed. All the curves in the plots result from the same 1A constant current profile.}
    \label{fig:degra}
\end{subfigure}
\caption{\textbf{Samples from the electrochemical battery model simulator.} Effect of varying the current load profile (Top) and the degradation parameters (Bottom) on the voltage discharge curves.}
\label{fig:introfig}
\end{figure}

In the first step, a large, representative dataset is generated comprising synthetic voltage curves resulting from different input current profiles and various ageing conditions covering different combinations of ageing parameters. 
%Since the subsequent step will rely on the representativeness of this dataset, it is important that it covers a wide spectrum of operating conditions. 
For this, we apply the recently introduced open-source,  physics-based NASA simulator of the electrochemical battery model \cite{2021_nasa_prog_models, chetan,daigle2016end}. A brief overview of this model and of the employed simulation engine are provided in the Supplementary Material. The main ingredient necessary to create our dataset is a set of parameters representing the degradation level of the battery. In this work, we focus on two of such parameters, namely $q_{max}$, capturing the total amount of available active Li-ions, and $R_{0}$, capturing the increase in the internal resistance. An illustration of the effect of varying these parameters on the corresponding voltage discharge curve (with a constant load profile) is shown in Fig. \ref{fig:degra}. To create such a representative dataset, a large number of $(q_{max},R_0)$-pairs and current profiles are sampled so that the resulting voltage curves generated by the simulator are as diversified as possible. For our largest dataset, the values of $q_{max}$ and $R_0$ were drawn from uniform distributions with supports between $5000$ $C$ and $8000$ $C$ and $0.017$ $\Omega$ and $0.45$ $\Omega$, respectively. These numbers have been chosen to reflect the values typically encountered in real batteries. With regard to the choice of the  load profiles, we consider both constant and variable profiles. For the variable profiles, we focus on piecewise constant trajectories. These are defined by indicating  a set of transition points (the values where the current changes value) and a corresponding set of constant values. We group different load profiles according to the number of transitions they include: the larger this number is, the greater the level of complexity of the corresponding current profile.
%As shown in Section \ref{results}, we assign each profile to a specific class according to the number of transitions it includes. 
Fig. \ref{fig:constant_variable} shows some examples of current load profiles and the corresponding voltage discharge curves. More details on the data generation process can be found in the Supplementary Material.

\noindent

\subsection*{Ageing-Aware EoD Prediction via Dynaformer}
As mentioned above, the goal of the proposed method is to accurately predict a battery's voltage discharge curve for any planned current profile while taking ageing effects into account. To do so, we propose Dynaformer, a Transformer-based encoder-decoder neural network architecture \cite{tr} for long-term voltage discharge prediction. Transformers represent the current state of the art in natural language processing \cite{kenton2019bert,brown2020language} and have recently been applied to other domains, such as computer vision \cite{vit}, speech recognition \cite{baevski2020wav2vec}, music generation \cite{huang2018music,vonrutte2022figaro}, time-series classification \cite{zerveas2021transformer} and forecasting \cite{zhou2021informer}, and symbolic mathematics \cite{lample2019deep,biggio2021neural}, among others. Both the encoder and decoder in Transformers are based on the attention mechanism, which allows them to selectively focus on the most relevant parts of the input sequence and to handle inputs of different lengths. This is a particularly useful property for our problem since the length of the voltage curves depends on the EoD point, which varies across different samples.\\
The main components of our model, along with the main steps involved in the EoD prediction process, are illustrated in Fig. \ref{fig:mainfigure}.
In the proposed framework, the role of the encoder is to extract information about the degradation level. This information is then used to condition the decoder in its prediction of the voltage discharge curve. In practice, we use the first few minutes of discharge of the battery -- along with the corresponding initial part of the current profile -- as input to the encoder for ageing inference. The decoder then receives as input the full current profile and the embedding output by the encoder, and predicts the discharge curve associated with the input profile and the implicitly inferred degradation condition. We would like to emphasize that we do not impose any explicit supervision on the encoder, i.e. we do not use the ground-truth degradation parameters to separately train the encoder. In fact, such parameters are not available at test time and our goal is instead to infer the ageing level of the battery implicitly and in an end-to-end manner. The only training signal we provide to the model is the ground-truth voltage discharge curve.
While for the encoder we resort to the original architecture, our decoder includes some significant modifications, which we introduce in the following. It is a well-known fact  \cite{tr} that Transformers are not well suited  to processing long input sequences because the time and space complexity of the attention mechanism scale as $\mathcal{O}({L^2})$, where $L$ is the input length. Since the input to the decoder can possibly be very long (current profiles can be several hours long, resulting in sequences of thousands of time steps), we resort to a strategy similar to the rationale motivating the design of Vision Transformers \cite{vit}. In \cite{vit}, the authors propose to decompose an image into multiple sub-patches and treat each of them as a single token to be processed by a standard Transformer. The resulting model achieves performance on par with or superior to that of state-of-the-art image classification models. Following a similar strategy, we subdivide the input time series into $\frac{L}{n}$ small sub-sequences, each of length $n$. Each sub-sequence will now be treated as a single $n$-dimensional token. This approach strongly alleviates the computational effort resulting from the application of the attention mechanism.\\
The training of the model is set up as a standard regression problem, whereby the output of the decoder, $\hat{\mathbf{y}}$, representing the full voltage discharge curve, is directly compared to the ground-truth voltage $\mathbf{y}$ via the calculation of the mean squared error (MSE). More details on the model's architecture and training can be found in the Supplementary Material.

\subsection*{Adaptation to Real Data via Fine-Tuning}
Despite its very good modelling capabilities, the employed simulator is not able to model some subtle degradation dynamics that manifest themselves in real data, in particular towards the end of the discharge curve (see, for example, Fig. 5 in \cite{daigle2016end}). This gap between simulated and real data can lead to imprecise EoD predictions if the model has been trained exclusively on simulated data. To overcome this limitation, we draw inspiration from natural language processing \cite{sun2019fine} by adopting a transfer learning approach, which uses only a limited amount of real data to \enquote{adapt} the model and reduce its bias towards simulated data. 
Thanks to the extensive training performed on synthetic data and to the high level of fidelity of the simulator, the model has already acquired a strong inductive bias on the general behaviour of a battery operated under various operating conditions and ageing levels. As a result, even a small subset of real data is sufficient to successfully adapt and transfer the model to the real world. Such a fine-tuning procedure is particularly relevant in cases where real data are scarce or at least not sufficiently abundant to satisfy the requirements of modern large over-parametrized neural networks \cite{kaplan2020scaling}. The \textit{simulation-to-real gap} described above is responsible for sub-optimal performance when a model trained on simulated data is directly applied to the real world. In our experiments, we show that this gap can be closed with fine-tuning, resulting in a model that quickly \textit{learns} how to deal with specific and subtle properties of real-world batteries, instead of relying on a complex -- and often unavailable -- physics-based description of such phenomena.

\section*{Results}
% experimental setup
\paragraph{Experimental Setup.} We assess the performance of the proposed approach by splitting our analysis into two parts. In the first step of our evaluation, we compare the performance of Dynaformer against two deep learning baselines on a simulated dataset with the aim of assessing our method's interpolation and extrapolation capabilities in a controlled setting. In particular, we compare our technique with another Sequence-to-Sequence (S2S) approach based on Long Short-Term Memory (LSTM) units, as well as a standard feed-forward fully-connected neural network (FNN) whose output is a pointwise prediction of the target trajectory. While Dynaformer and the LSTM model can deal with input sequences of different lengths, the FNN is specifically designed for inputs of fixed dimensionality. Thus, for a fair comparison, we start our analysis with constant current profiles, which can be simply represented by a single real number indicating the corresponding current intensity. We then extend our study to variable current profiles with different levels of complexity. Since the FNN is not directly applicable in such settings, we only compare the performance of the proposed approach with the LSTM model.\\ 
In the second step of our analysis, we use data collected from real Li-ion batteries and investigate whether our method -- which has been pre-trained on simulated data -- can be effectively transferred to real data. The goal of this last part is to evaluate whether the proposed method is able to close the simulation-to-real gap with only a small amount of fine-tuning. Additional information regarding the baseline methods can be found in the Supplementary Material.

\paragraph{Metrics.}
We resort to the standard root-mean-squared error (RMSE) between predicted and ground-truth voltage to measure how well the algorithm is able to model the shape of the voltage discharge curve.\\
In addition to the RMSE, we propose a new metric, the relative temporal error (RTE) between the predicted end-of-discharge point and the ground-truth discharge point, for a more precise characterization of the quality and robustness of the predicted EoD time. The main motivation behind this metric is 
that we do not want to simply assess whether, given the exact current profile leading to EoD, the model is able to reconstruct the corresponding discharge curve. We also want to verify whether, given a longer (shorter) profile than the exact one, the model is able to terminate before the end of the current profile (not let the voltage drop before the actual EoD time). To calculate this quantity, we evaluate the EoD prediction performance of the considered methods for different lengths of the same input current profile. In practice, we start from a cropped version of the input current trajectory (70\% of the original length) and then gradually extend it up to to 130\% of its initial length. For each of these scenarios, we compute the absolute error between the predicted EoD and the ground-truth EoD, normalized by the length of the original input profile. Ultimately, this results in a set of relative errors, one for every considered length, and the final RTE is defined as the maximum value over this set. It is, therefore, a \emph{worst-case} measure of the performance of the method. \\
Additional information about the metrics -- including the pseudocode of the algorithm to calculate the RTE -- can be found in the Supplementary Material.

\subsection*{Performance Evaluation on Simulated Data}\label{results}
\subsubsection*{Performance Evaluation on Constant Load Profiles} \label{results:constant}
\begin{figure}[H]
    \centering
    \includegraphics[scale=0.34]{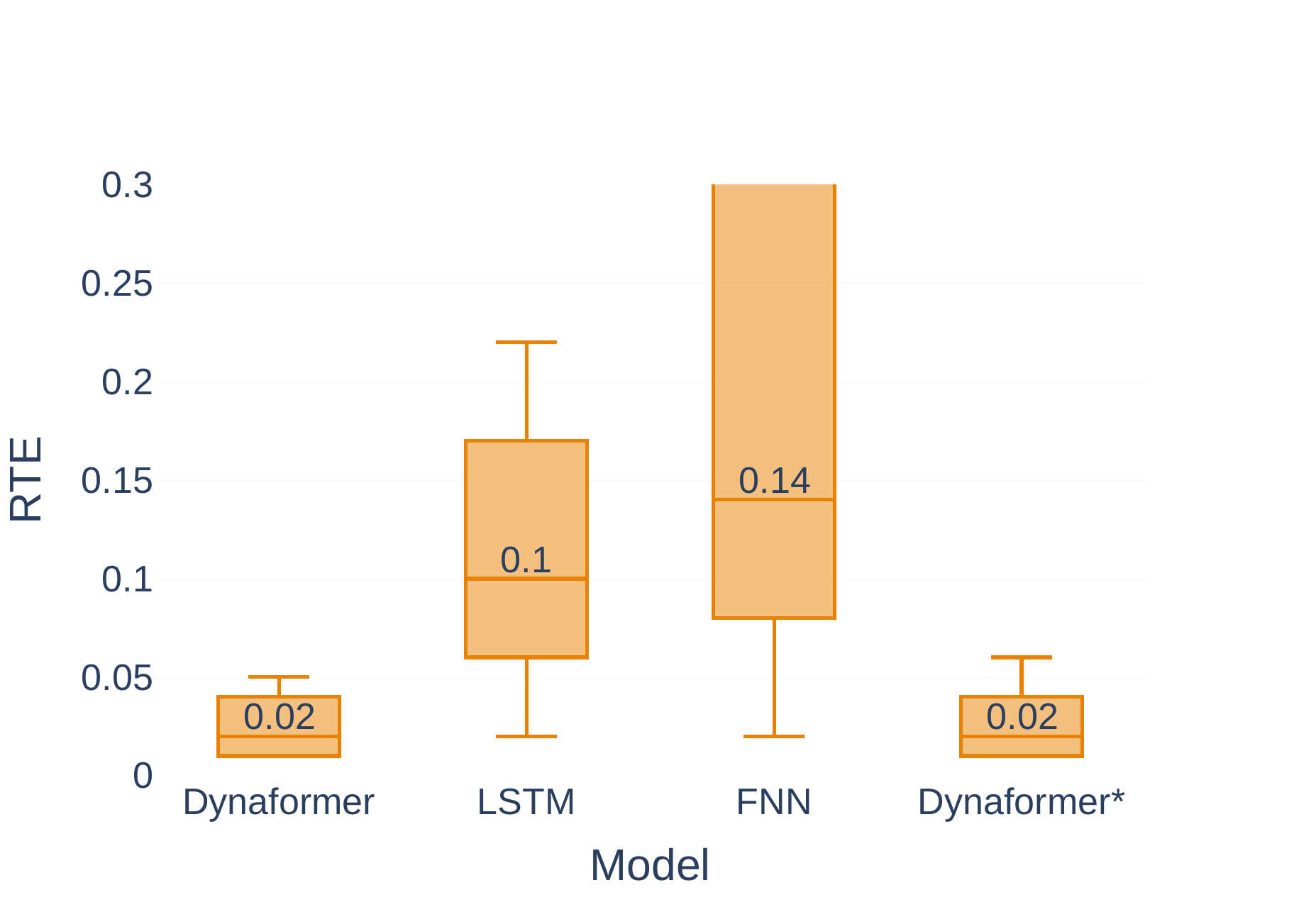}
    \includegraphics[scale=0.34]{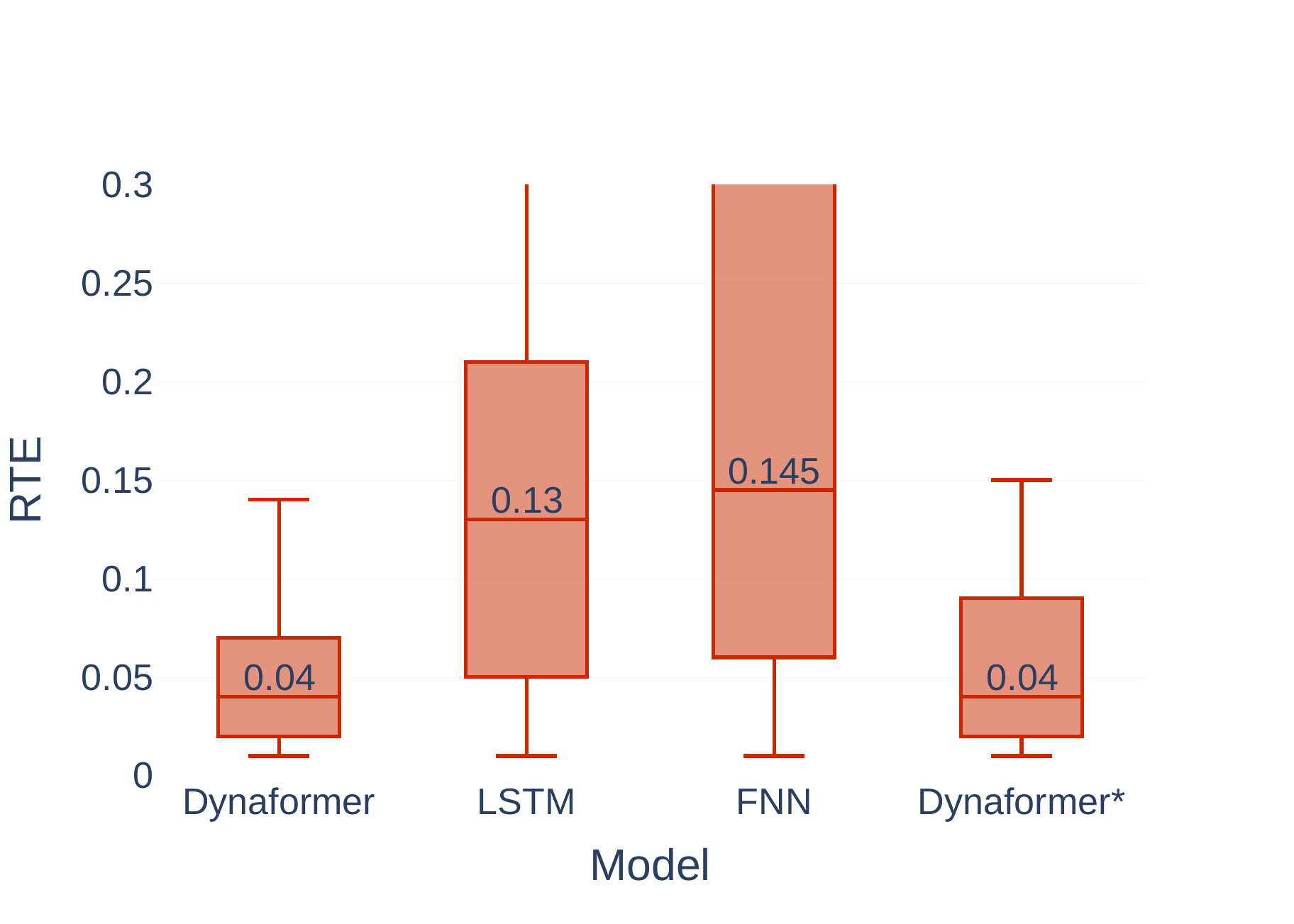}
    \caption{\textbf{Results on constant load profiles.}    Performance of our model and the baselines in terms of RTE. On the left, we present the interpolation performance; on the right, the extrapolation performance. The whiskers denote the 5th and 95th percentiles. Dynaformer* denotes our method trained with variable current profiles.}
    \label{fig:constant_current}
\end{figure}

We start our analysis by evaluating the performance of the proposed Dynaformer on constant current profiles and comparing it against two baselines: a vanilla feed-forward fully-connected neural network (FNN) and an LSTM-based seq-to-seq model \cite{cho2014learning}.
%and a DeepONet \cite{Lu_2021}, a state of the art architecture for learning operators which we have adapted for the case of learning the mapping from current to voltage. 
We consider two variants of our algorithm: the first is trained exclusively on constant profiles and is, thus, specialized in the task of interest of this section; the second is trained on variable current profiles.  
%As outlined in Section \ref{methodology}, we could not benchmark against standard attention-based architectures due to their prohibitive memory requirements for the large sequence length of the current and voltage profiles. 
Apart from the latter version of our model, we have trained all the methods with the same dataset, comprising current profiles with current values drawn uniformly from the interval [0.5A, 3A] and the corresponding voltage profiles drawn from uniform distributions with supports between $5000$ $C$ and $8000$ $C$ and $0.017$ $\Omega$ and $0.45$ $\Omega$, respectively. For testing, we have generated two separate test sets, one with degradation values different from those used to construct the training set, yet sampled in the same range in order to test the interpolation performance, and another with degradation values outside the training set to evaluate the extrapolation performance. For this last dataset, the values of the degradation parameters are up to 10\% larger/smaller than the largest/smallest values used in the training distributions of each parameter.
The results of the considered methods on both the interpolation and the extrapolation test sets are summarized in Fig. \ref{fig:constant_current}.

\vspace{-5mm}
\noindent
As can be seen from the results displayed in Fig. \ref{fig:constant_current}, in the interpolation test set, Dynaformer achieves a median RTE of 0.02, which is significantly better than the 0.1 and 0.14 of the LSTM and the FNN, respectively. Furthermore, Dynaformer performs well on test data in the extrapolation regime, achieving 0.04 median RTE, which is only slightly worse than the result in the interpolation range. Our algorithm trained on variable current profiles performs comparably to its counterpart trained on constant profiles, both in the interpolation and extrapolation regimes. Given that the constant current dataset is a subset of the variable current dataset, this result shows that there is no performance degradation if Dynaformer is trained with larger and more heterogeneous data. \\
We further investigate how the performance of our model is affected by degradation parameters within different ageing regimes. Each circle in Fig. 4 represents the RTE for a unseen combination of $q_{max}$ and $R_0$ at 1 A (left) and at 2 A (right). The shaded 2D area depicts the  the range of degradation parameters used to generate the training discharge trajectories. As the figure shows, the RTE can be predicted with high accuracy for combinations of $q_{max}$ and $R_0$ that are within or close to the training ranges, before gradually decreasing as the values of $q_{max}$ and $R_0$ move \emph{far outside} the training distribution. This result not only suggests that Dynaformer can accurately predict the EoD for degradation levels comprised in the training set, but also that its predictions are reliable for ageing conditions in the extrapolation regime, a property that endows our method with a high level of robustness. 

\begin{figure}[H]
    \centering
    \includegraphics[scale=0.36]{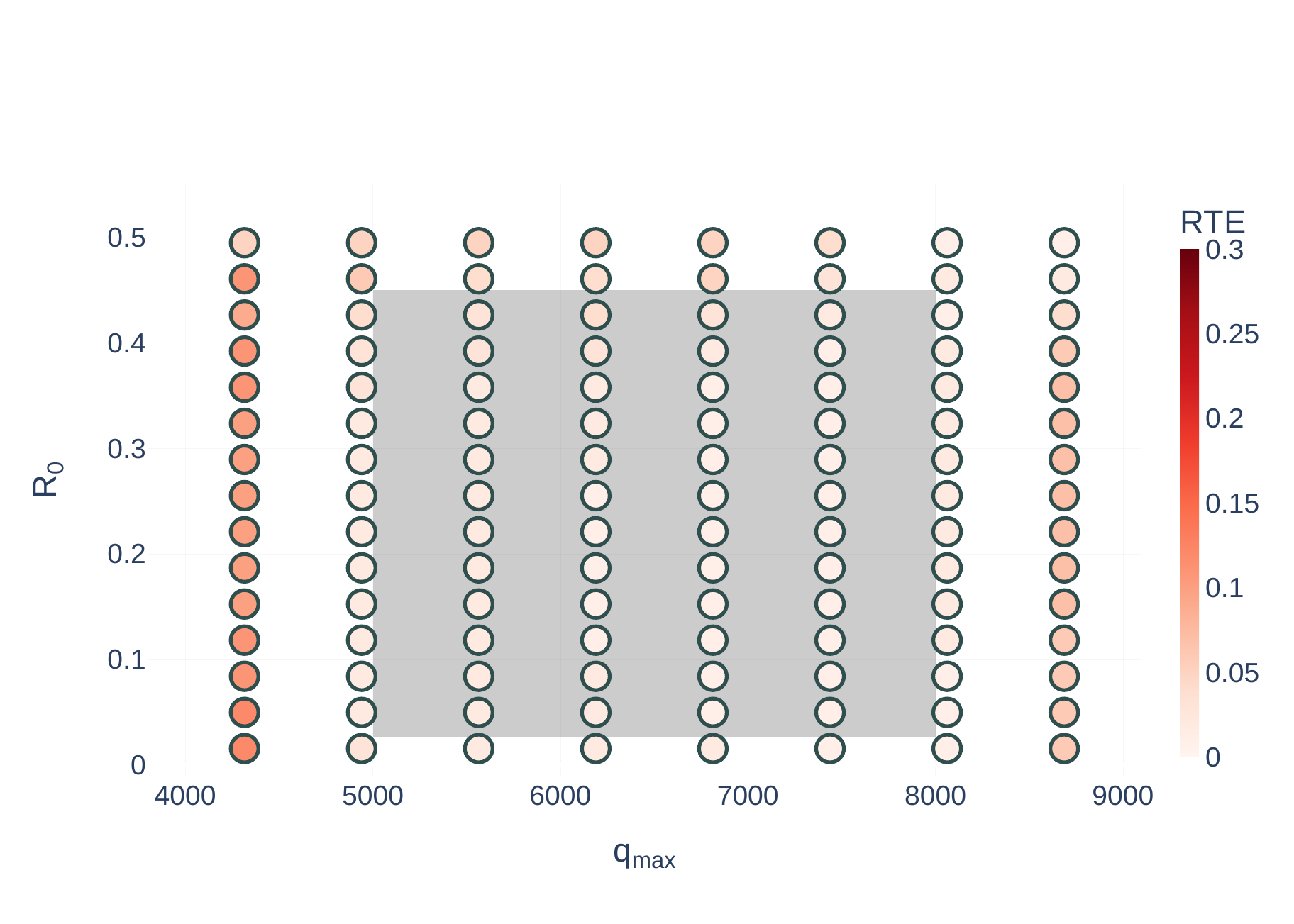}
    \includegraphics[scale=0.36]{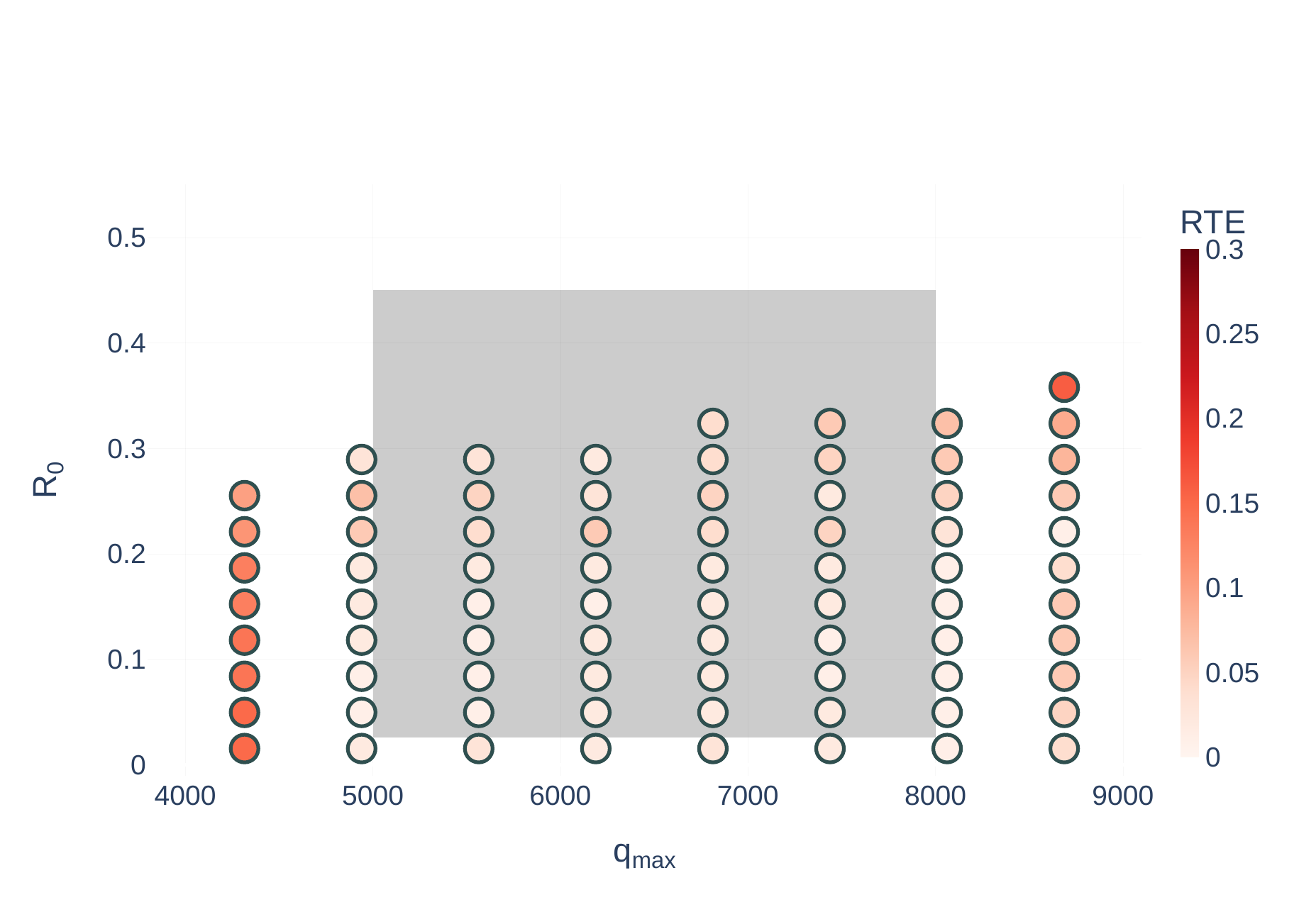}
    \caption{\textbf{Generalization performance analysis.}
    Generalization performance of Dynaformer with respect to the degradation parameters $q_{max}$ and $R_0$. The grey shaded area represents the interpolation region. Current is fixed at 1 A (left) and 2.0 A (right). The empty spaces in the right panel are due to unrealistically short voltage profiles that have not been included in the evaluation.}
    \label{fig:genQR}
\end{figure}
\vspace{-7mm}

\subsubsection*{Performance Evaluation on Variable Load Profiles}\label{results:variable}
We now consider the more challenging setting of variable current profiles. We generate a new training dataset with the same sampling procedure as in the constant current case. However, instead of using constant current profiles, we generate piecewise constant sequences with random length. We define a transition as the instant at which the current changes its value. Our training set was generated with profiles with up to five transitions and we tested the generalization performance of our model on new profiles -- not included in the training set -- with up to eleven transitions. The results for Dynaformer are summarized in Fig. \ref{fig:currents_variable}, while in the Supplementary Material, we report the same analysis for the LSTM model.

\begin{figure}[H]
    \centering
    \includegraphics[scale=0.34]{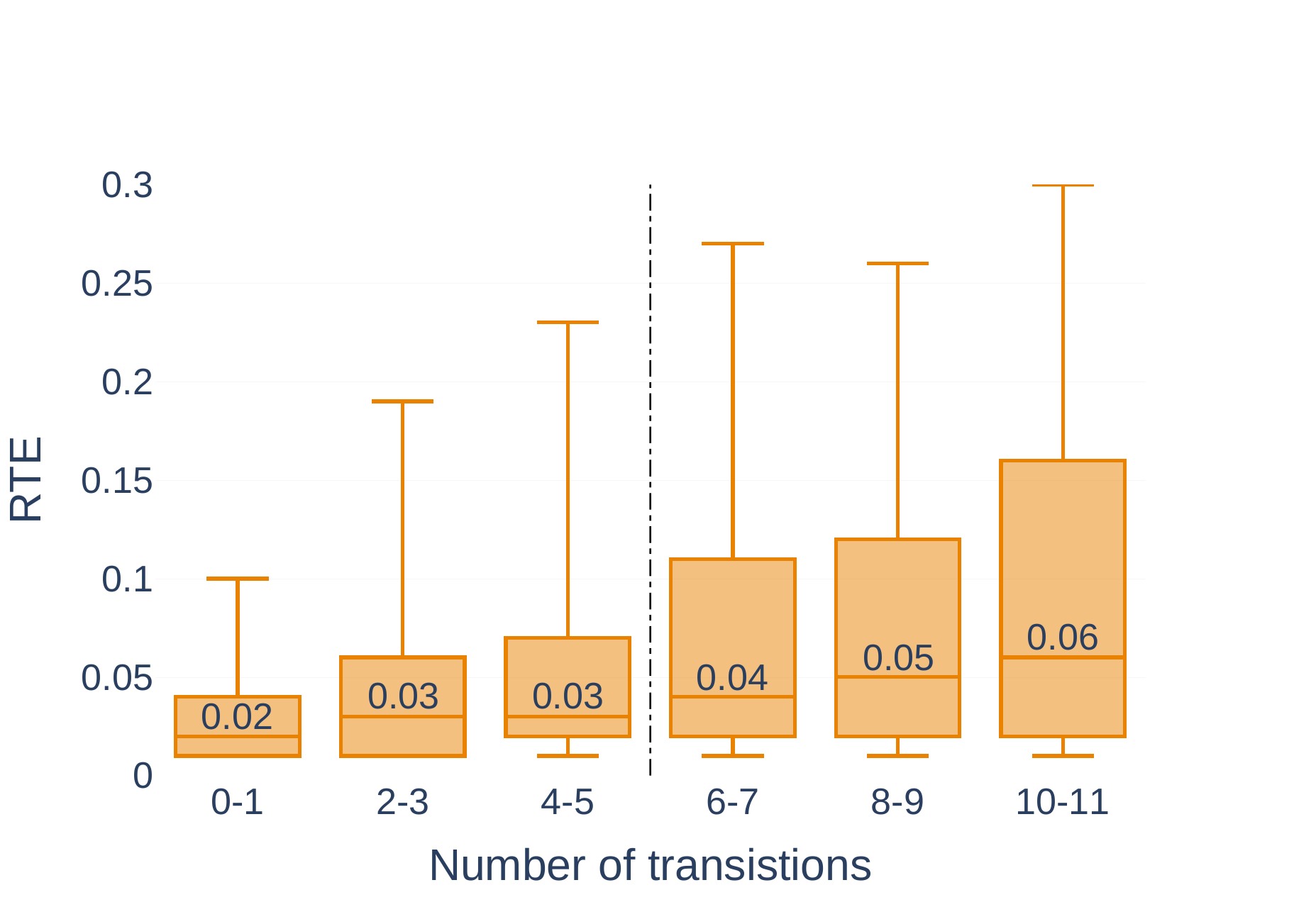}
    \includegraphics[scale=0.34]{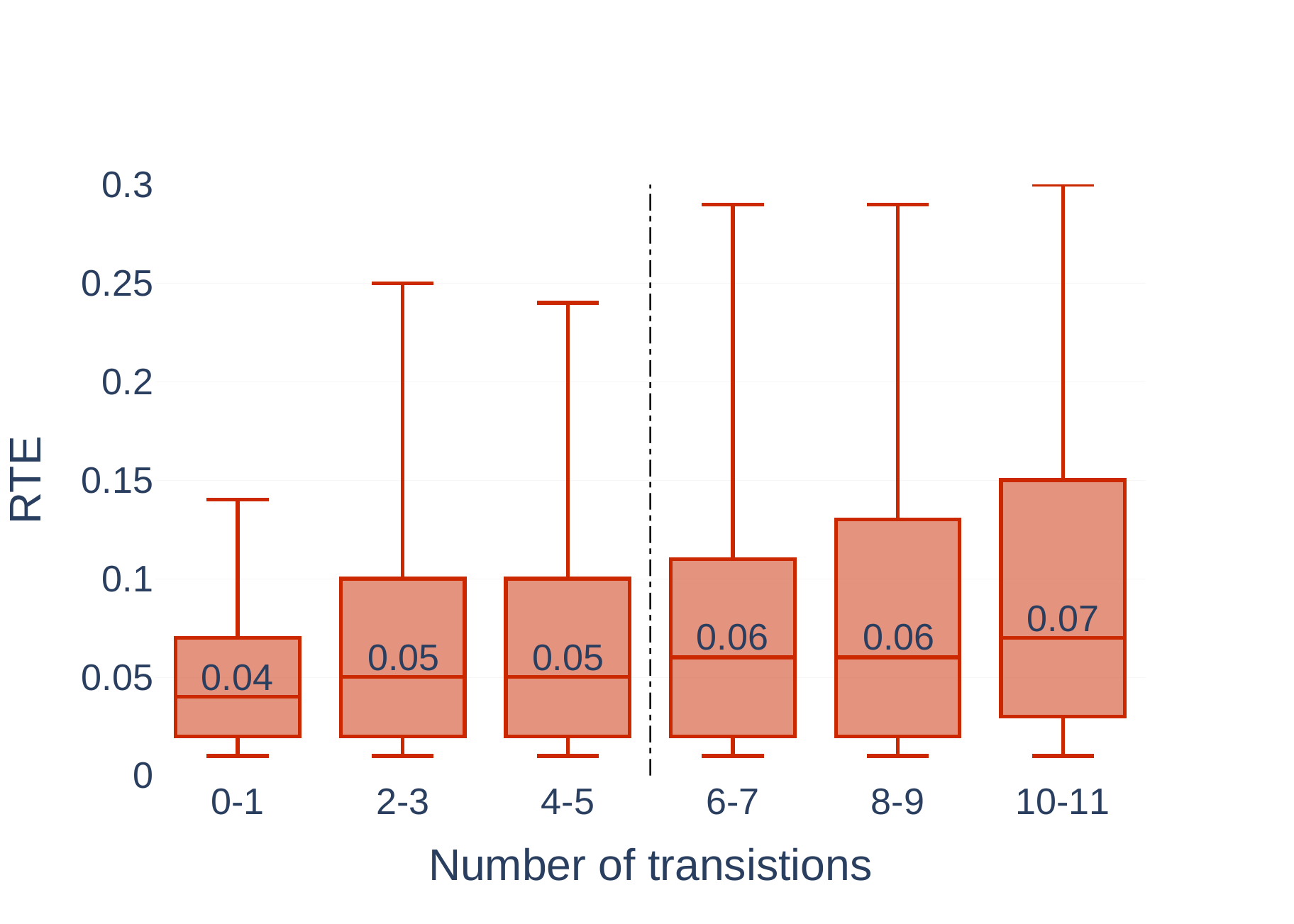}
    \caption{\textbf{Results on variable load profiles.} Performance of Dynaformer for values of $q_{max}$ and $R_0$ in the interpolation regime (orange, left) and extrapolation regime (red, right). The whiskers denote the 5th and 95th percentiles. The grey dashed vertical lines between the classes $[4$-$5]$ and $[6$-$7]$ transitions separate current profiles belonging to the interpolation regime and extrapolation regimes (from 6 to 11 transitions).}
    \label{fig:currents_variable}
\end{figure}
\noindent
The left panel of Fig. \ref{fig:currents_variable} shows the performance of the algorithm for levels of degradation within the training distribution, while the panel to the right displays the performance of the proposed methodology in the extrapolation regime. For both panels, current profiles are grouped according to their number of transitions and a grey dashed line separates current profiles with the same (left) and a larger (right) number of transitions as the samples used in the training set. The results imply that Dynaformer is robust to the increase in the level of complexity of the current profiles, with only a slight decrease in performance as a function of the number of transitions. Furthermore, the model generalizes well to current trajectories never seen during training, i.e. with more transitions than the profiles in the training set. Even more importantly, the performance of Dynaformer experiences only a small decrease when values of the degradation parameters out of the training distribution are considered (right panel). 

\begin{figure}[H]
    \centering
    \includegraphics[scale=0.24]{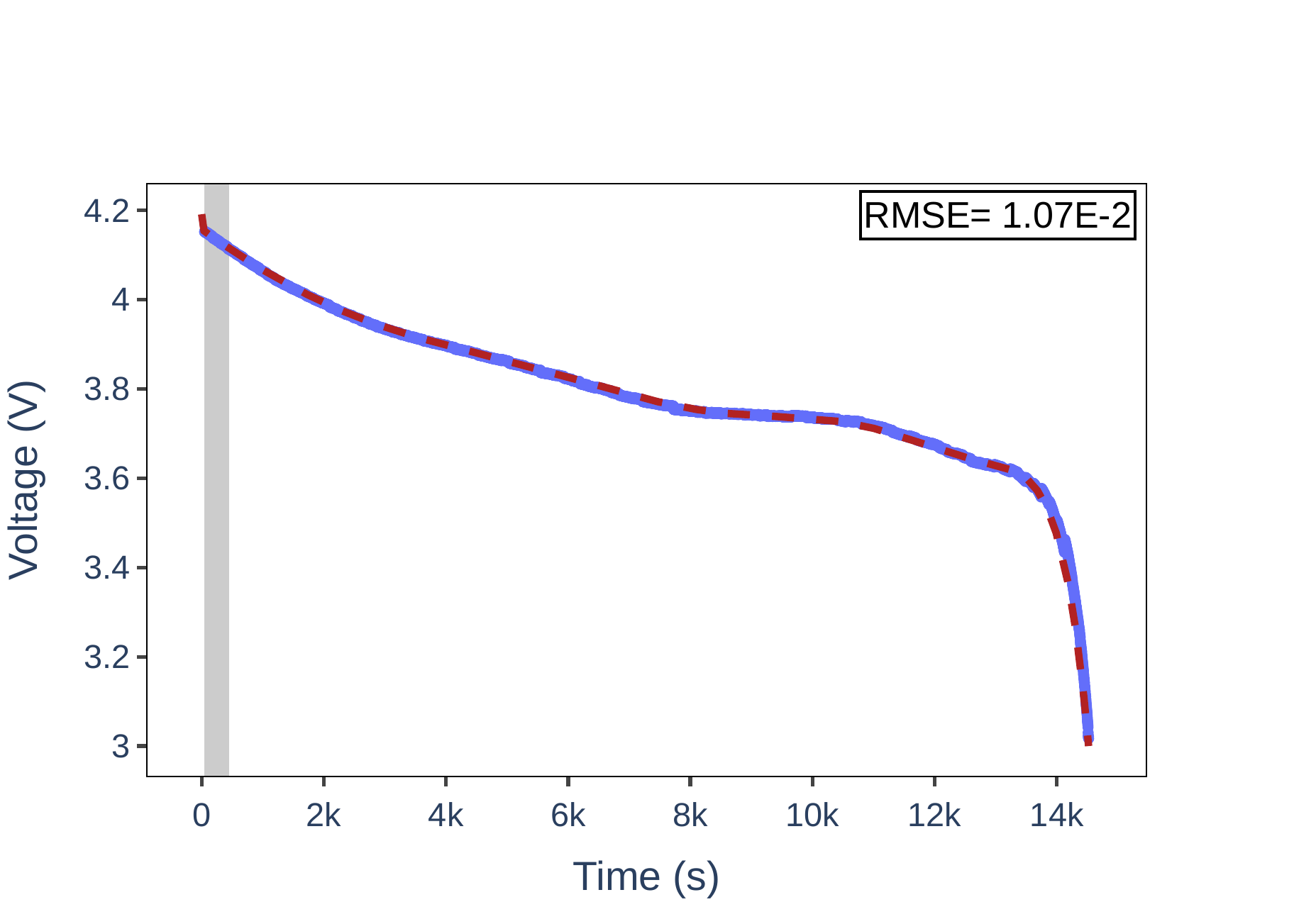}
    \includegraphics[scale=0.24]{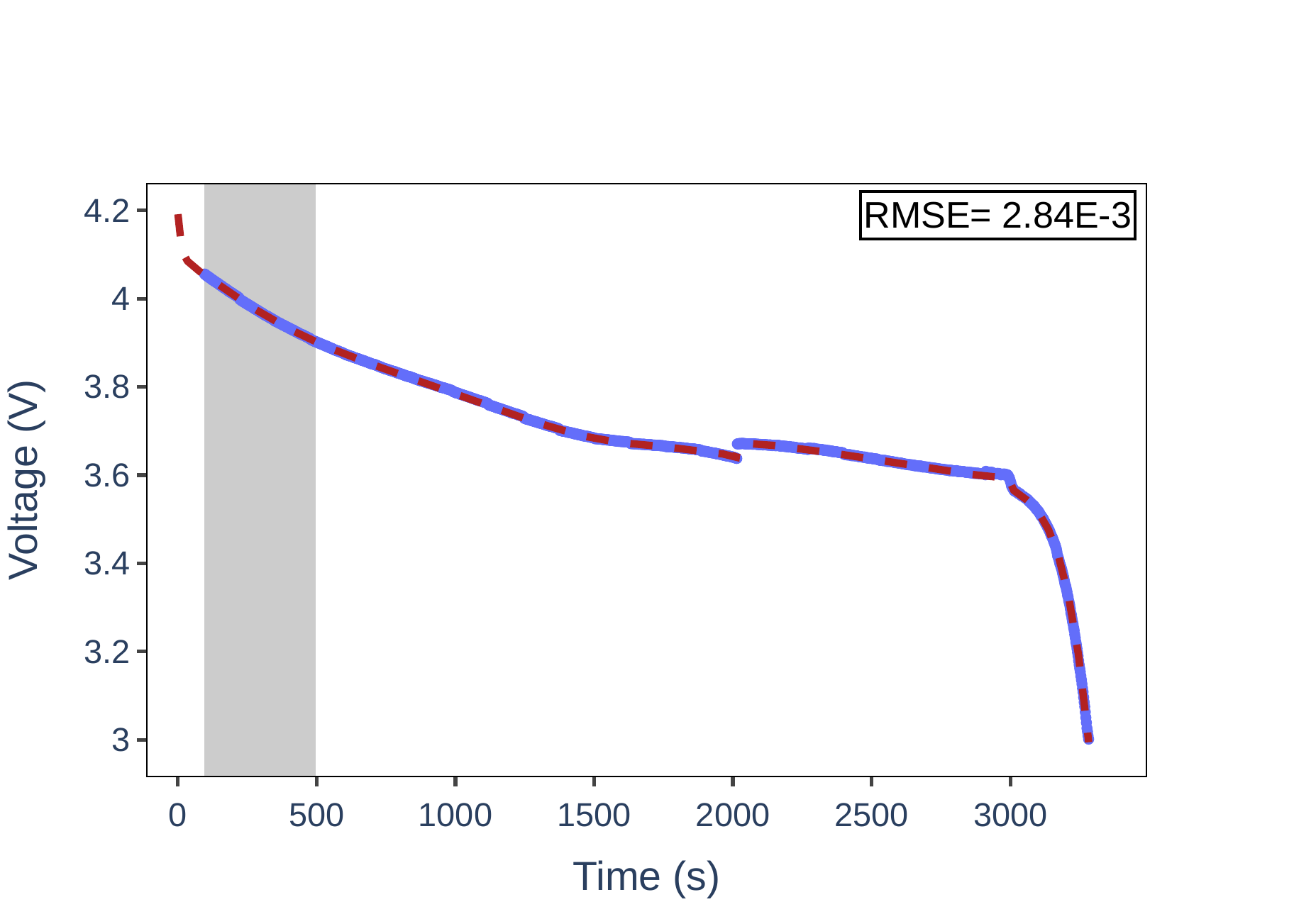}
    \includegraphics[scale=0.24]{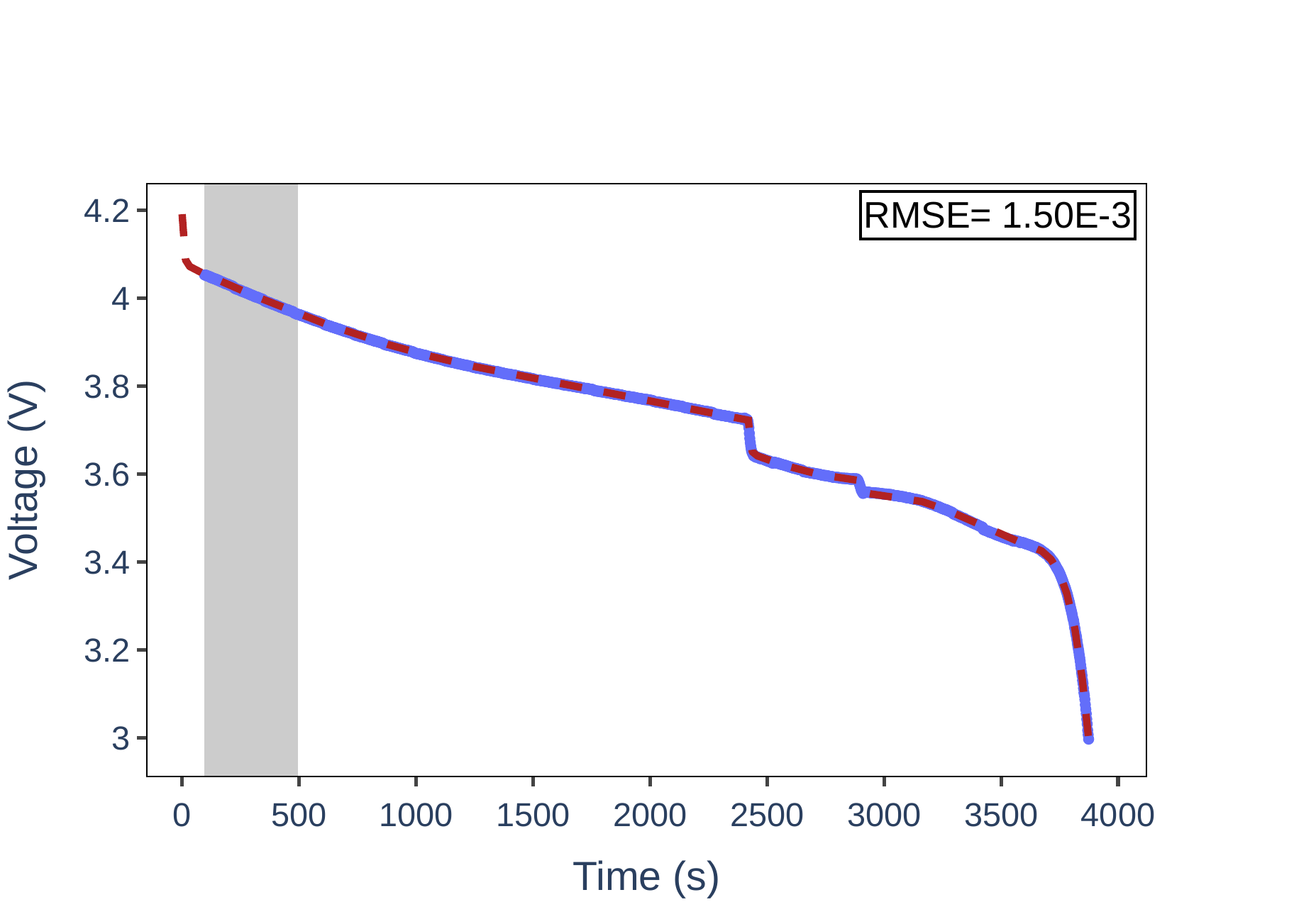}
    \includegraphics[scale=0.24]{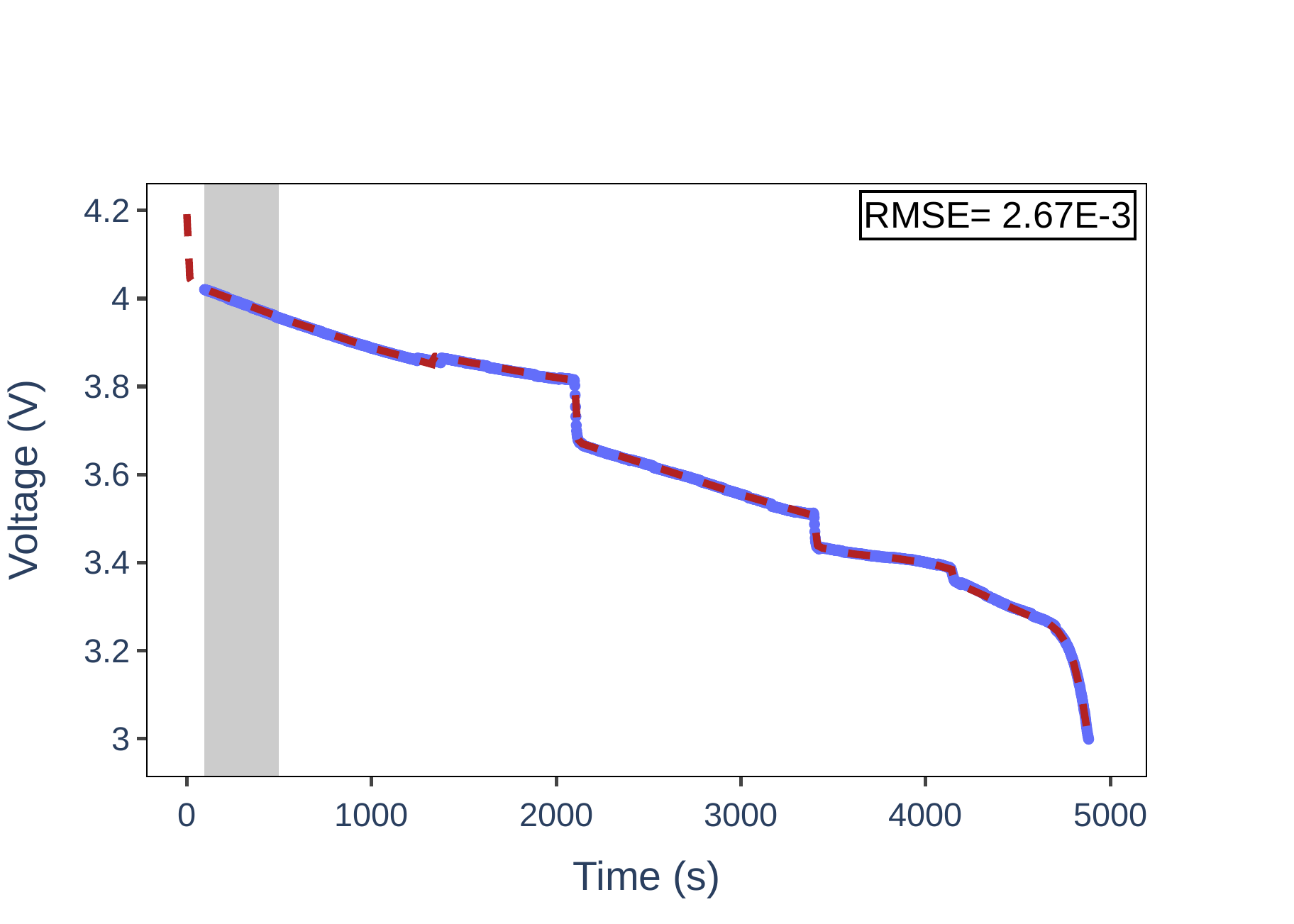}
    \includegraphics[scale=0.24]{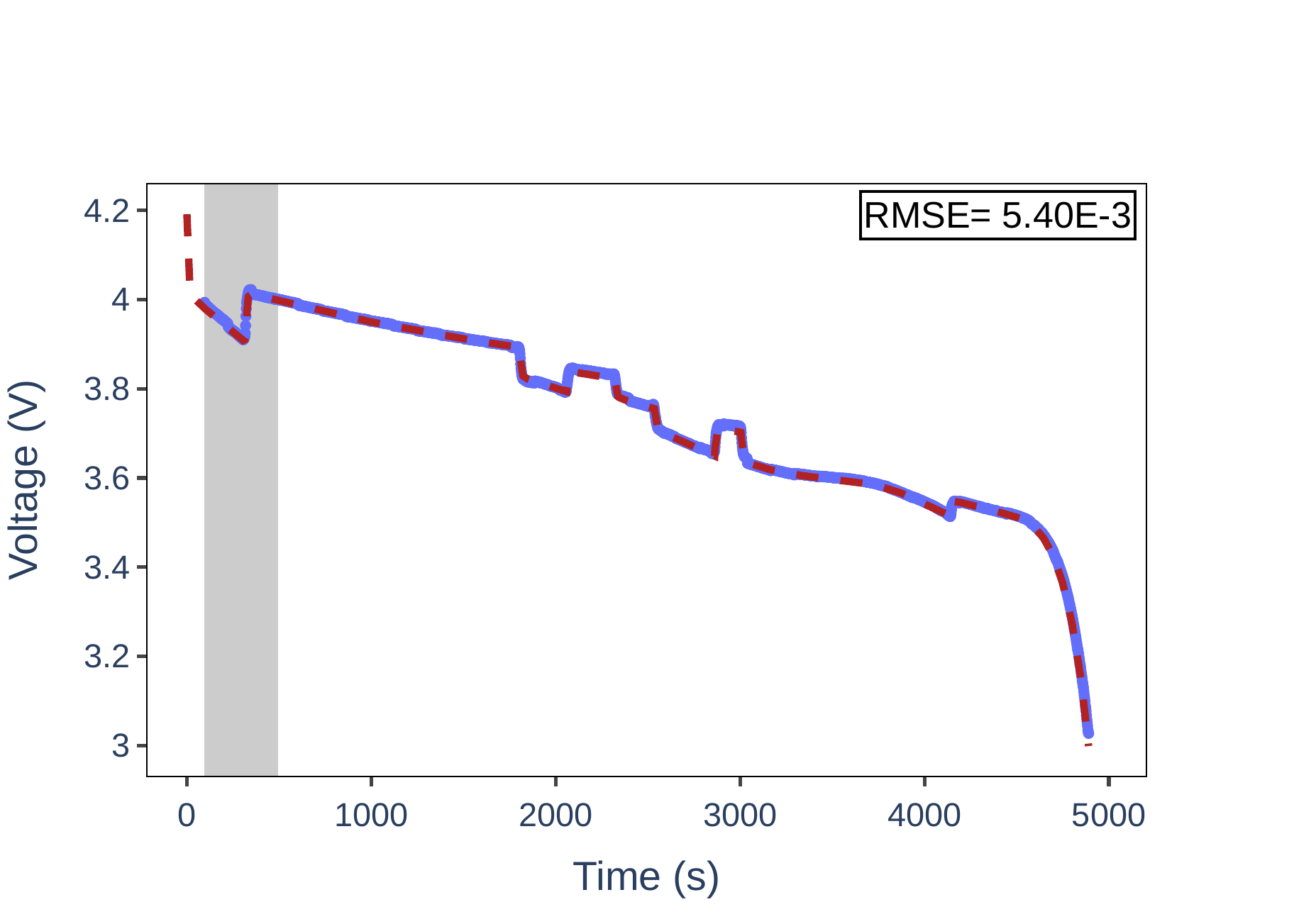}
    \includegraphics[scale=0.24]{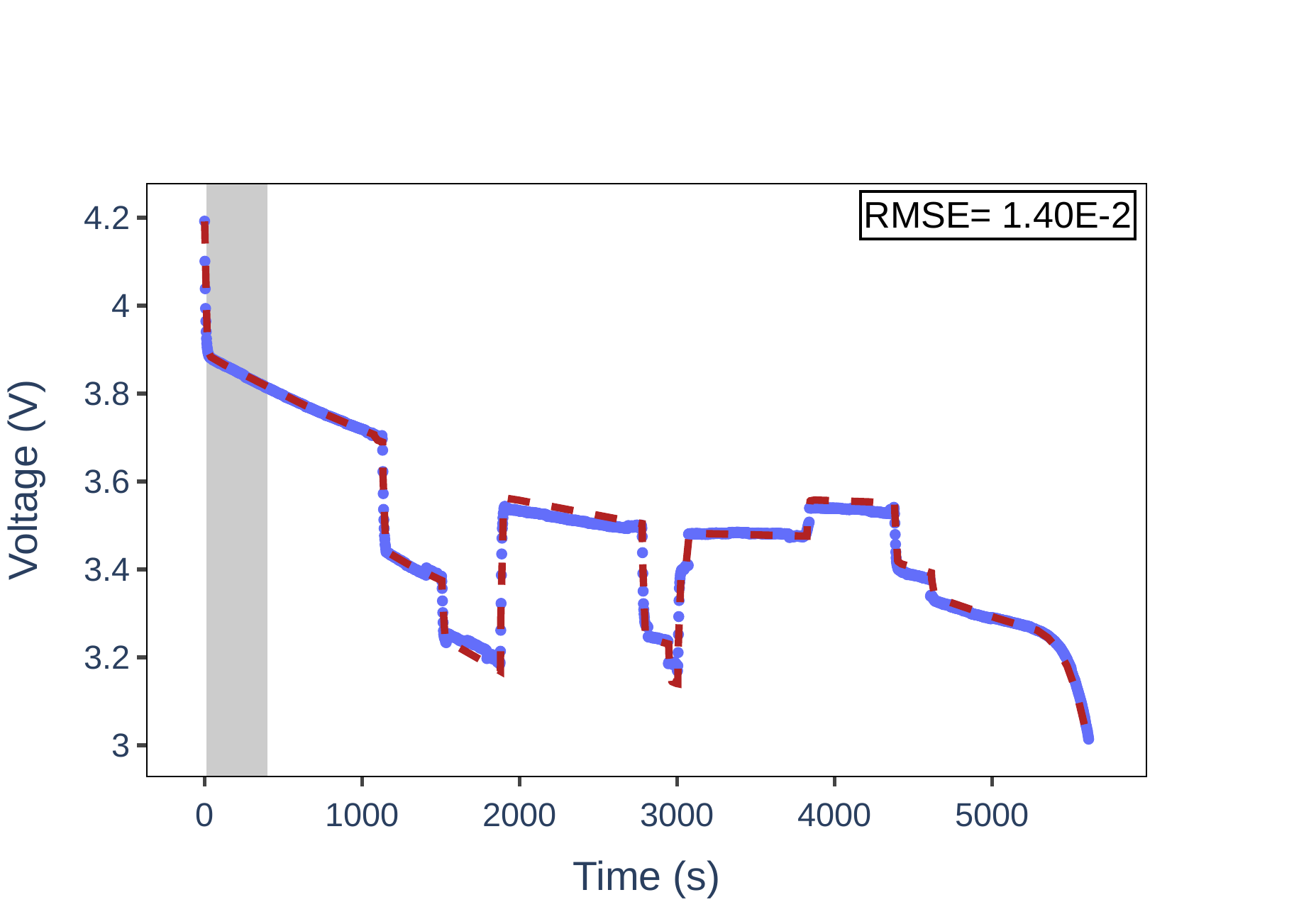}
    \caption{\textbf{Illustration of the predictions of Dynaformer as the complexity of the current profiles increases.} (Top) Number of transitions from left to right: 0, 2, 3.
    (Bottom) Number of transitions from left to right: 4, 8, 11. The red dashed line represents the ground truth, while the blue line represents our model's prediction. The grey shaded area indicates the context window used by the encoder. The box on the top right of each panel includes the RMSE of our model in approximating the ground truth.}
    \label{fig:samples}
\end{figure} 
\noindent
We visualize the voltage profiles generated from current profiles of different levels of complexity, along with the corresponding predictions output by our model, in Fig. \ref{fig:samples}. Dynaformer predicts the discharge curves very precisely and is effective in capturing the multiple sharp transitions characterizing complex profiles (e.g. bottom right panels).\\
These results show that the model also performs well in the variable current case and is able to generalize well both to unseen degradation levels -- thus confirming the results obtained in the constant current case -- and to more complex current profiles than those comprised in the training set.

\subsection*{Implicit Ageing Inference}\label{interpret}
In this section, we investigate whether our trained model is able to extract information on the ageing level from data without having explicitly been trained to do so. To answer this question, we inspect the output of the encoder for inputs corresponding to various degradation states. The encoder is responsible for the extraction of the information related to the degradation parameters and its function is thus critical for the final performance of the algorithm.  
\begin{figure}[H]
    \centering
    \includegraphics[scale=0.4]{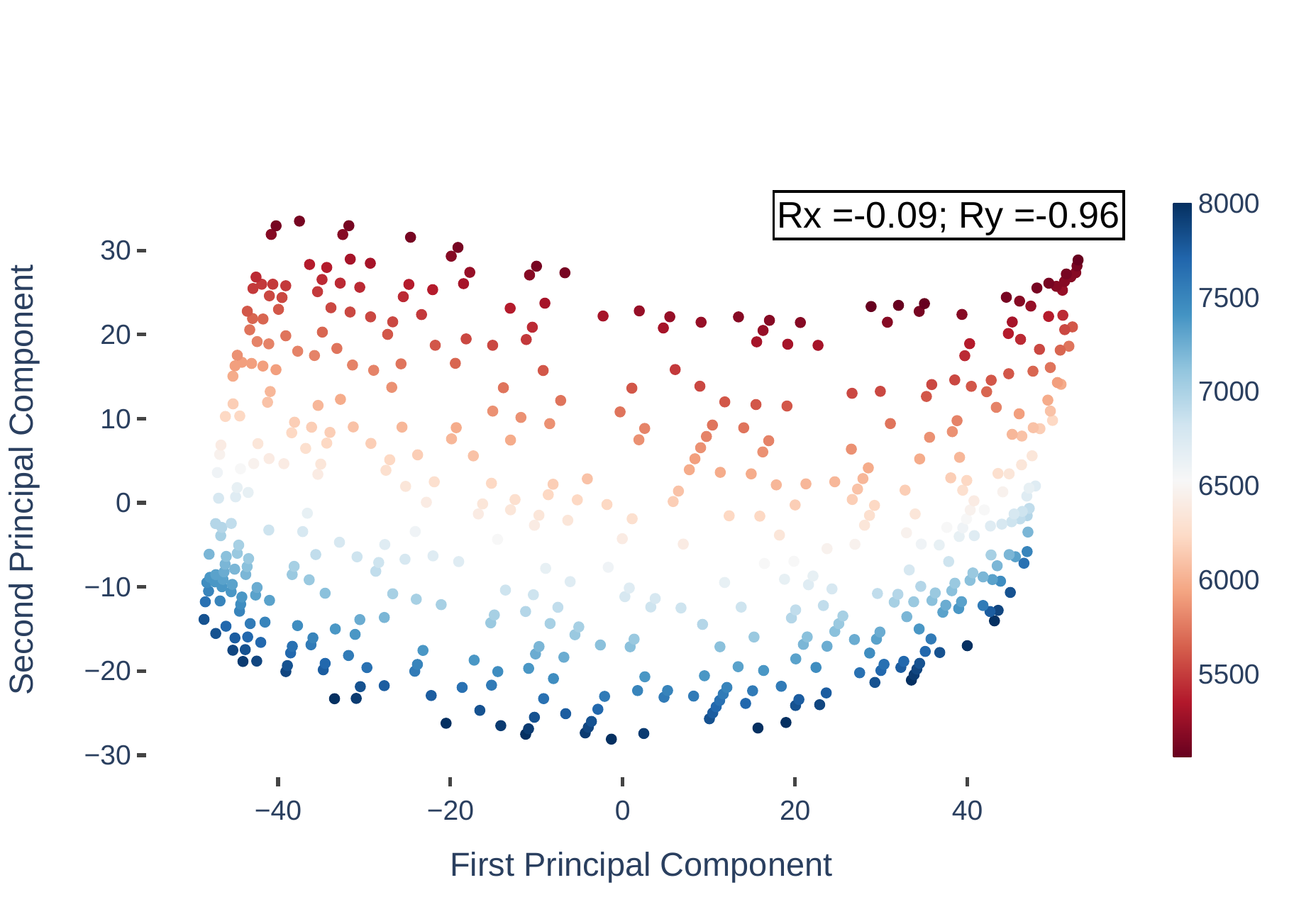}
    \includegraphics[scale = 0.4]{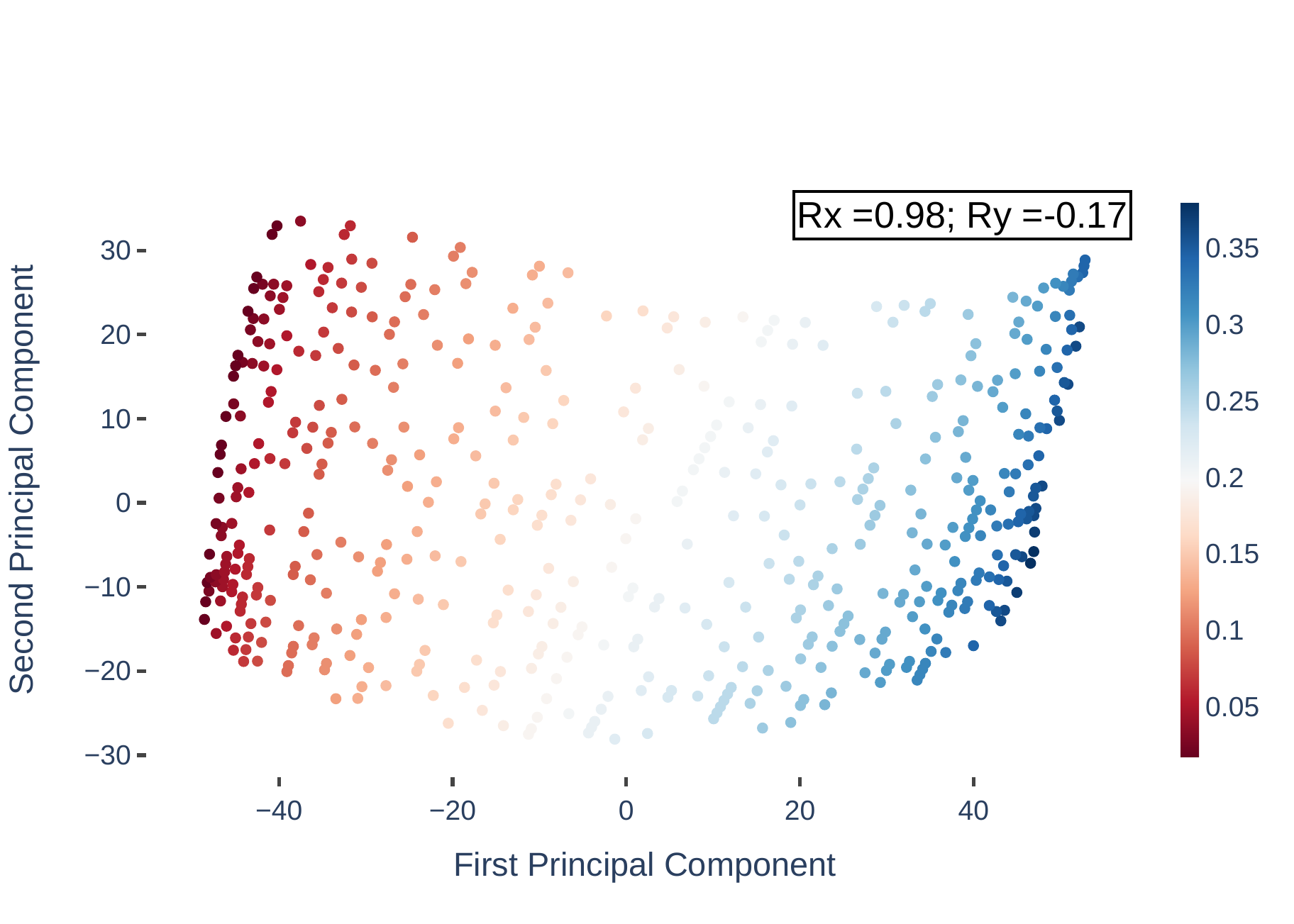}
    \caption{\textbf{Implicit parameter inference.} Principal component analysis of the output of the encoder embedding. (Left) Each point in the latent space is coloured according to the value of $q_{max}$. (Right) Each point in the latent space is coloured according to the value of $R_0$. In the top right corner of the left (right) panel, the Pearson's correlation coefficients of the first and second principal components with $q_{max} (R_0)$ are shown.}\label{fig:physics}%
\end{figure}
\noindent
Fig. \ref{fig:physics} shows the first two principal components of the encoder output, where each point is coloured according to the values of $q_{max}$ (left) and $R_0$ (right). It is apparent from the two plots that there is a high level of correlation between $q_{max}$ ($R_0$) and the first (second) principal component of the encoder output. To quantify this statement, we calculate the Pearson's correlation coefficient between the values of $q_{max}$ and $R_0$ of each sample and the corresponding principal components. We find a correlation coefficient of 0.96 (0.98) between $q_{max}$ ($R_{0}$) and the first (second) principal component, thus confirming our initial observation.
This means that Dynaformer is able to automatically discriminate between different ageing conditions without being \emph{explicitly} trained to do so. In other words, the model has learnt how to perform parameter inference \emph{jointly} with EOD prediction. In the language of causality \cite{scholkopf2022causality}, another way to express this concept is that the model has effectively learnt the \emph{causal factors of variations} in the training data. Such factors are closely related to the data-generation process and the model has demonstrated the ability to successfully \emph{disentangle} these factors in its latent space. This aspect confers a significant degree of interpretability to our model: in practice, one can simply infer the degradation parameters by inspecting the region of latent space where the output of the encoder lies.

\subsection*{Performance Evaluation on Real Data}
\begin{figure}[H]%
    \centering
    \begin{subfigure}{5cm}
    \includegraphics[scale=0.3]{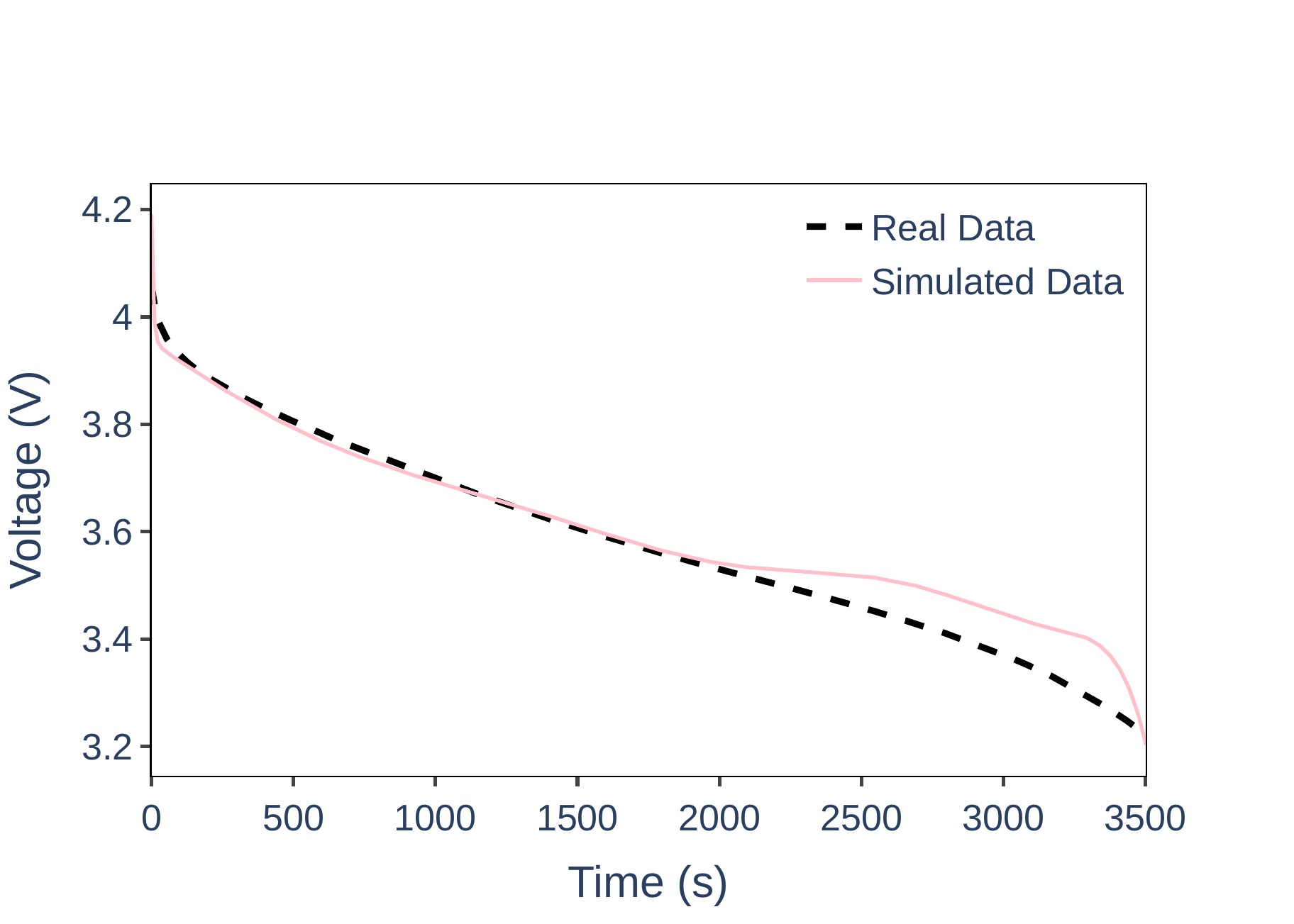}
    \caption{}\label{fig:orig}
    \end{subfigure}
    \qquad
    \begin{subfigure}{5cm}
    \includegraphics[scale=0.3]{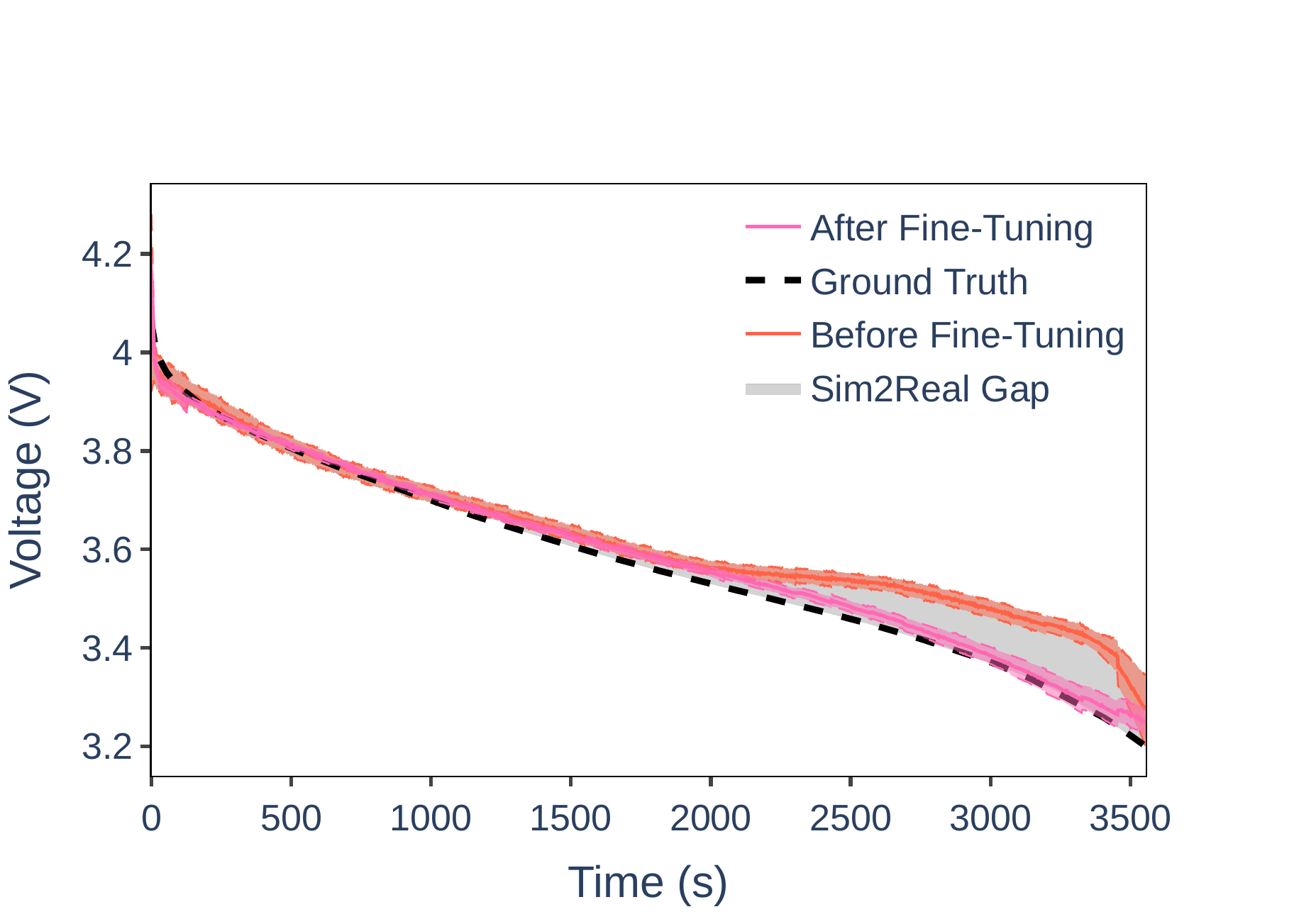}
    \caption{}
    \end{subfigure}
    \qquad \\
    \begin{subfigure}{5cm}
    \includegraphics[scale=0.3]{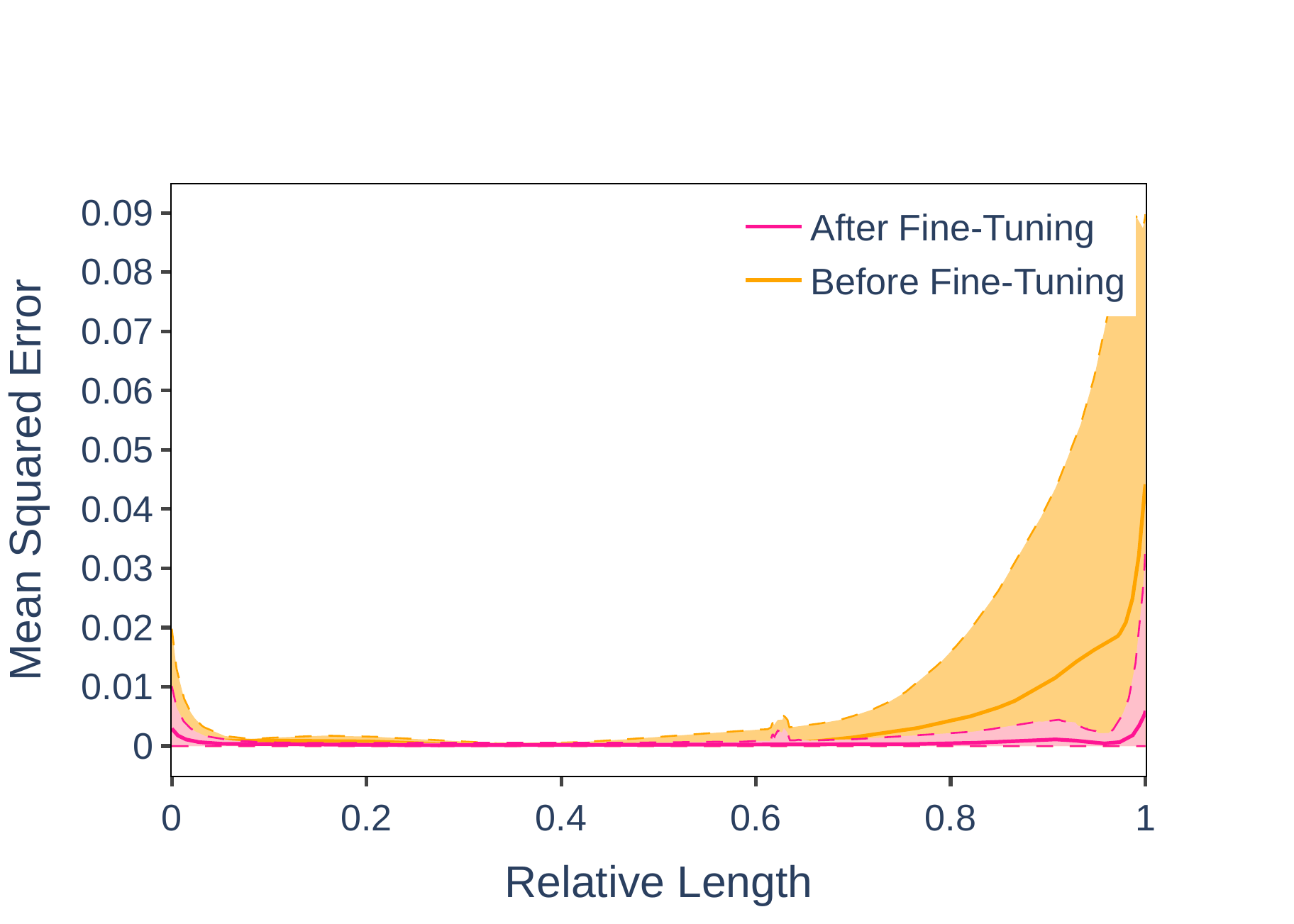}
    \caption{}
    \end{subfigure}
    \qquad
    \begin{subfigure}{5cm}
    \includegraphics[scale = 0.3]{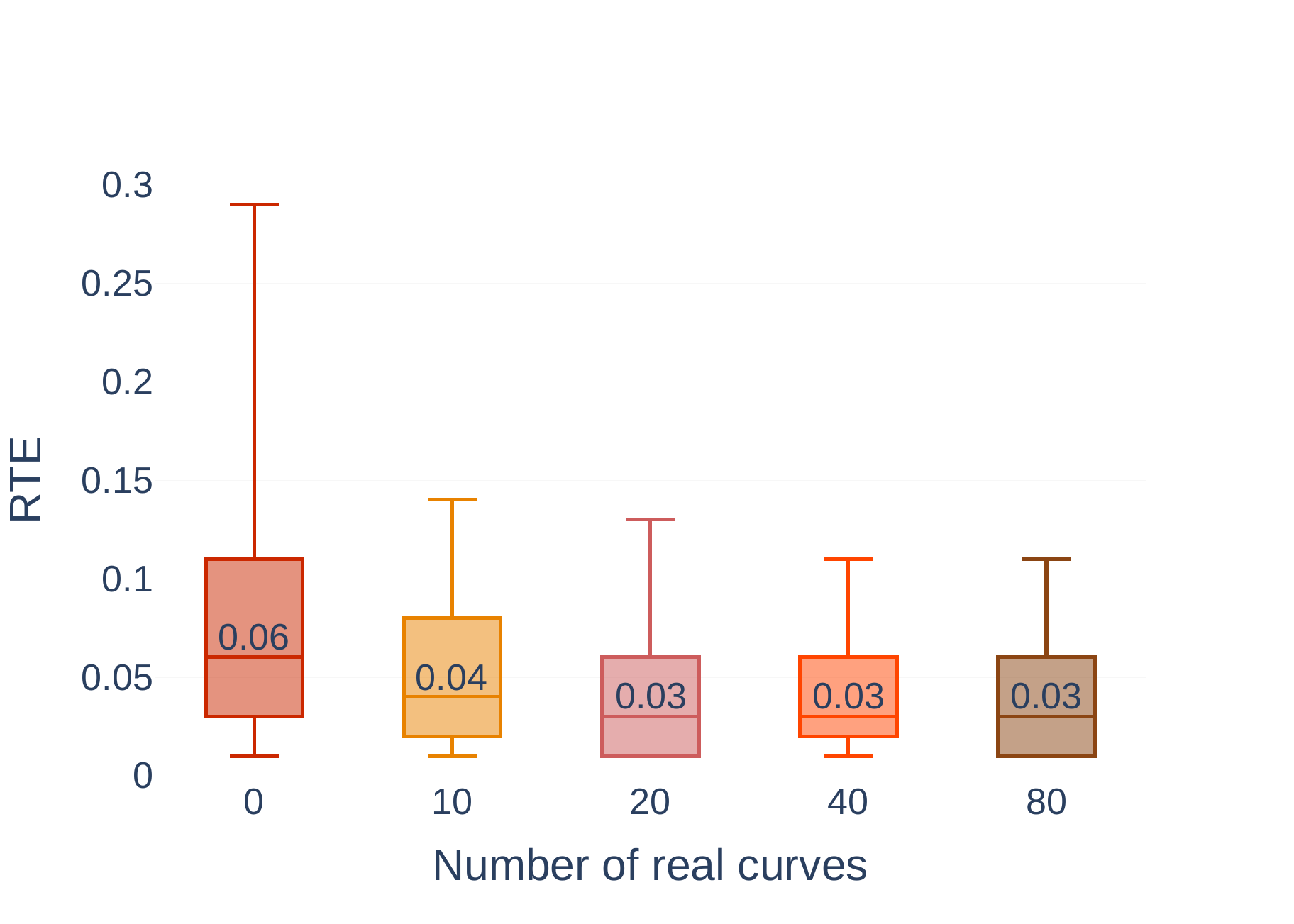}
    \caption{}
    \end{subfigure}
    \caption{\textbf{Adaptation to real data via fine-tuning.} (a) Discrepancy between simulated data and real data. (b) Fine-tuning closes the sim2real gap \protect\footnotemark . (c) MSE distribution across the test real dataset for the model before and after fine-tuning. (d) The RTE as a function of the size of the fine-tuning dataset.  The whiskers denote the 5th and 95th percentiles.}%
    \label{fig:finetuning}%
\end{figure}

In the second step of our evaluation process, we assess the performance of Dynaformer on data collected from a set of real batteries. The data used for the batteries in this work are from the open source NASA dataset \cite{bole2014randomized}. In particular, we consider four batteries, identified as RW9,RW10,RW11 and RW12, operated under a constant current of 2A until the EoD point of 3.2 V is reached. Each battery undergoes multiple charge-discharge cycles, resulting in a progressive degradation. For each battery, about 80 cycles are available. More details about the dataset can be found in the Supplementary Material.\\%These are 18650 single cells, 8 of which are used in the data set, where each cell is subjected to randomized discharge profiles to observed degradation and ageing in these batteries.   \luca{we only use constant current profiles (reference discharge): RW9/RW10/RW11/RW12. I would be nice to say some words about the types of batteries.}
%\luca{CHETAN: Need details on the batteries here}
Under the hypotheses that the simulated training dataset is large and covers a wide enough set of diverse operating conditions and that the employed simulator possesses a relatively high degree of fidelity to the dynamics of real batteries, we might expect the error made by the model on real data not to be overly large. This is indeed what we observe in our experiments: when directly applied to real data, Dynaformer attains a median relative error of about $6\%$. While this performance is already acceptable, such a result is sub-optimal in the sense that the simulator used to train our model is not able to fully describe the details of the discharging process occurring in real batteries (see, for example, Fig. 5 in \cite{daigle2016end} and Fig. \ref{fig:finetuning} (a)).  This is a manifestation of the simulation-to-real gap, i.e. the incapacity of model-based approaches to fully characterize the phenomenon they aim to describe. Despite this aspect, the already good performance delivered by the model when directly applied to real data seems to suggest that such a gap is relatively small and is particularly prone to occur in the last part of the discharge cycle. To mitigate this effect, in this work, we follow a transfer learning procedure by fine-tuning our pre-trained model on a small subset of real data to close the simulation-to-real gap and further improve performance on real data. We explicitly require that the size of the fine-tuning dataset be small since this is typically the scenario encountered in realistic applications. Please note that directly training our model on such a small dataset inevitably results in massive overfitting, preventing the model from generalizing to \emph{new} real data instances. The goal of fine-tuning in this case is to show that a model pre-trained on simulated data has already acquired a good enough inductive bias on the battery discharge process and its ageing mechanisms, that only a relatively small fraction of real data is sufficient to refine its performance and close the simulation-to-real gap. \\

\noindent
\footnotetext{Uncertainty bounds ($\pm 3\sigma$) are calculated via the Monte Carlo Dropout technique \cite{gal2016dropout}. More details on this can be found in the Supplementary Material.} 
\noindent
%\luca{give some details about the train/test split.}
To assess the performance of Dynaformer, we first fine-tune it on a subset of the available real data and then we test it on the remaining part. More specifically, we adopt a cross-validation procedure in our experiments: we use the data from one out of 4 batteries for training and the data from the remaining three for testing. The results of the fine-tuning procedure are shown in Fig. \ref{fig:finetuning}. In Fig. \ref{fig:finetuning} (b), we show the difference in performance between the model before and after fine-tuning on a sample test curve (not seen during fine-tuning). Beforehand, the simulation-to-real gap is apparent (grey area) and Dynaformer fails to accurately model the full discharge curve. After fine-tuning, the model adjusts its output, which is now aligned with the ground truth. In Fig. \ref{fig:finetuning} (c), we provide a more quantitative picture of the effect of fine-tuning by showing the MSE distribution across the test samples as a function of the relative length of the voltage curves. 
The effect of fine-tuning manifests itself in a drastic decrease in MSE, particularly pronounced towards the end of the test sequences.
Finally Fig. \ref{fig:finetuning} (d) indicates that by increasing the size of the fine-tuning training dataset, the performance improves quickly. Please note that although the best results are achieved by employing the full dataset of 80 curves, nearly optimal performances can already be obtained with smaller training sizes. This suggests that the pre-training phase is key to our approach, infusing a strong inductive bias into the model and allowing it to achieve improved performance on real data via a simple fine-tuning procedure.\\
Overall, the model after fine-tuning attains a median RTE of approximately 0.03, an improvement of about 50\% with respect to its value before fine-tuning.

\section*{Discussion}
%% contributions statement
In this work, a novel approach to ageing-aware EoD prediction is proposed. Our method is trained on a large set of simulated data and our extensive experimental evaluation demonstrates that it is efficient, accurate and can handle long input time series of arbitrary complexity. The proposed Dynaformer learns to implicitly extract the degradation parameters, and can be easily adapted to real data, overcoming the simulation-to-real gap. Our experiments show that it delivers remarkable generalization performance, with respect to both new degradation parameters and new load profiles. Furthermore, it outperforms the considered competitive deep learning models by a significant margin. \\
The proposed methodology paves the way for long-term planning of missions characterized by heterogeneous and complex load profiles. Its predictions are only based on a handful of current/voltage observations and can estimate the voltage profile up to EoD time with a very high precision. This implies that battery-powered systems can be reliably operated for more cycles and closer to the actual EoD, without the need of large margins due to imprecise and uncertain predictions. This would result in the full exploitation of the battery charge, prolonging, thereby, also its useful lifetime and leading to significant cost savings and improved sustainability.\\
Our work can be extended in multiple ways. First, the proposed methodology could be easily applied to datasets generated by alternative battery simulators. Similar to modern language models based on Transformers \cite{brown2020language}, an interesting direction would also be to 
create a larger dataset comprising very different types of batteries and use it to train a bigger version of the current architecture. The goal would then be to obtain a \emph{universal} model for EoD prediction that is highly versatile among a wide range of battery types.
A second interesting direction arises from the observation that our model represents an accurate \emph{differentiable} simulator of the real battery system. Thanks to its differentiable nature, gradient-based methods can be used to, for instance, determine which input current profile leads to a specific voltage discharge trajectory given a certain degradation level.
Third, overall the Dynaformer approach is general and we envision its application to learning very different system (and degradation) dynamics. In particular, our method 
can be potentially employed in all those engineering applications where successful performance on a long-term sequence prediction task is dependent on both a preliminary inference stage and on the effective processing of an input conditioning signal. These characteristics are shared across a wide range of engineering applications. 
We leave the exploration of all these open directions to future work.
\newpage

\section*{Data Availability}
We publicly release the datasets used to train our models along with the code to generate them at the following \href{https://github.com/SymposiumOrganization/Dynaformer}{link}.

\section*{Code Availability}
We publicly release the implementations of our method and the baselines along with the scripts to visualize the results at the following \href{https://github.com/SymposiumOrganization/Dynaformer}{link}.

\bibliographystyle{ieeetr}
\bibliography{main}

\newpage

\appendix
\begin{center}
\huge{\textbf{Dynaformer: A Deep Learning Model for Ageing-aware Battery Discharge Prediction. Supplementary Information}}\\
%\vspace{1cm}
%\normalsize
%Luca Biggio$^{1,3}$, Tommaso Bendinelli$^{3}$, Chetan Kulkarni$^{4}$, and Olga Fink$^{2}$\\
%\vspace{0.5cm}
%$^1${Data Analytics Lab, ETH, Zürich, Switzerland}\\
%$^2${Laboratory for Intelligent Maintenance and Operations Systems, EPFL, Switzerland}\\
%$^3${CSEM SA, Alpnach, Switzerland}\\
%$^4${KBR, Inc., NASA Ames Research Center, Mountain View, CA 94035, USA}
\end{center}

\section*{Supplementary Notes}

\subsection*{Supplementary Note 1: Li-Ion Battery Model}\label{batterymodel}
Synthetic curves were generated using  the NASA Prognostic Model library \cite{2021_nasa_prog_models} which implements the single cell battery model described in \cite{chetan}. The model is based on lumped-parameters analysis of Li-ion batteries and includes the effect of ageing and degradation. In particular, it consists of a set of ordinary differential equations -- written in terms of the functions $\mathbf{f}$ and $h$ -- which describes  the dynamics of the system state $\mathbf{x}$ and the measured voltage ${V}$,
$$
\begin{aligned}
\mathbf{x}(k+1) &=\mathbf{f}(k, \mathbf{x}(k), {u}(k), \boldsymbol{\theta}) \\
V(k) &=h(k, \mathbf{x}(k),{u}(k), \boldsymbol{\theta})
\end{aligned}
$$
where $k$ denotes time, $u$ the input current and $\boldsymbol{\theta}$ is a set of parameters (including $q_{max}$ and $R_0$) characterizing the battery. The state vector $\mathbf{x}$ consists of the following components:
$$
\mathbf{x}=\left[\begin{array}{lllllll}
 V_{o}^{\prime} &
V_{\eta, p}^{\prime} & V_{\eta, n}^{\prime} &
q_{s, p} & q_{b, p} & q_{b, n} & q_{s, n}  \end{array}\right]
$$
where $V_{o}^{\prime}$ is the voltage drops due to the solid-phase ohmic resistances, $V_{\eta, p}^{\prime}$ and $V_{\eta, n}^{\prime}$ are the
voltage drops due to the charge transfer resistance and the solid electrolyte interface kinetics at the positive and negative electrodes respectively, $q$ is the total amount of available ions in the positive (suffix $p$) or negative (suffix $n$) in either surface (suffix $s$) or bulk (suffix $b$) volume. More details can be found in \cite{chetan}.\\
We used version 1.0.1 of the NASA Prognostic Model repository for our experiments, simulating the voltage profile to a threshold value of 3 V and randomizing the two ageing parameters $q_{max}, R_0$ as detailed in the main body. All other parameters and initial conditions were left unchanged to the typical values assumed for common commercial Li-ion 18650-type cells.

\subsection*{Supplementary Note 2: Datasets}\label{datasets}
In this section we provide additional details on the data used to train the considered models. We start by discussing our synthetic data generation process and we conclude by providing further information on real battery dataset used to fine-tune our model. We release all the data used in our experiments at this \href{https://github.com/SymposiumOrganization/Dynaformer}{link}.
\subsubsection*{Synthetic Datasets}
We simulate voltage discharge curves with the previously described NASA Prognostics library \cite{2021_nasa_prog_models}. We choose a sampling frequency of 0.5 Hz for both current profiles and voltage curves. Each voltage trajectory represents a full discharge of a battery with a nominal capacity of 2.1 Ah. The initial voltage (open circuit voltage) of the battery is 4.2 V, and we simulate it until 3 V to represent typical end-of-discharge conditions for Li-ion batteries. For both constant and variable current profiles, we filter out trajectories shorter than 500 seconds and longer than 20,000 seconds. 
\paragraph{Training Dataset}
\paragraph{Constant Current Profiles.}
We created a training dataset comprising 66,357 current/voltage profiles with a constant current drawn uniformly in the range between $0.5 A$ and $3.0 A$. $q_{max}$ samples were drawn uniformly from the region comprised between $5000 C$ and $8000 C$, while $R_0$ samples were drawn uniformly between  $0.017215 \Omega$ and $0.45 \Omega$.
\paragraph{Variable  Current Profiles.}
We created a dataset comprising 549,218 current/voltage profiles with piecewise constant current profiles with each constant segment characterized by a magnitude drawn uniformly in the range between $0.5 A$ and $3.0 A$. We sample the number of transitions in the piecewise constant current profiles uniformly between 0 and 6.

\paragraph{Test Dataset\\}\hspace{-0.4cm}The synthetic test sets were generated using the same simulator used for training, albeit with different parameters.
\paragraph{Constant Current Profiles.}
The interpolation test set includes 222 profiles with lengths from 1,200 to 17,200 seconds timestamps. These profiles were generated with $q_{max}$, $R_0$ within the same region as the training set. The extrapolation test set includes 292 profiles with either $q_{max}$, $R_0$, or both outside the interpolation region, within a maximum range of 10\% of the maximal interpolation values of $q_{max}$, $R_0$. For both sets, sampled constant current values were generated using a uniform distribution between 0.5A and 3A.
\paragraph{Variable  Current Profiles.}
The interpolation set includes 15,980 profiles, 2,336 of which have 0/1 transitions, 2,514 have 2/3 transitions, 2,189 have 4/5 transitions,  2,179 have 6/7 transitions, 2,028 have 8/9 transitions, and 1,952 have 10/11 transitions. The extrapolation set includes 9,598 profiles, 1,423 of which have 0/1 transitions, 1,578 have 2/3 transitions, 1,303 have 4/5 transitions, 1,314 have 6/7 transitions, 1,276 have 8/9 transitions, and 1,184 have 10/11 transitions. 

\subsubsection*{Real Datasets}
Battery dataset collected from the data repository of NASA’s Ames Research Center \cite{bole2014randomized} is being employed in this research work. This dataset consists of voltage discharge curves from four commercially available Li-Ion 18650 batteries referred to as RW9, RW10, RW11 and RW12. Data are collected with two different load profiles, namely Random Walk and Reference discharge profiles. For our experiments, we focus on the second class of curves. To create the dataset, batteries are first charged to their maximum voltage of 4.2 V and then loaded with a current of 2 A until the voltage reaches 3.2 V. This charge-discharge procedure is repeated over multiple cycles, resulting in the development of degradation effects on the batteries. More details about the dataset can be found in \cite{bole2014randomized, daigle2016end}.

\subsection*{Supplementary Note 3: Method and Baselines}\label{methodsandbaselines}
In this section, we provide more details on the architectures of our model and all the baselines. All the models have been implemented in Pytorch (version 1.10.1) and, for the sake of reproducibility, we provide open-source access to our code at the following link.
\subsubsection*{Dynaformer}
The Dynaformer consists of an encoder-decoder Transformer-based architecture. We describe the role and main parts of each of its components in the following paragraphs.
\paragraph{Encoder.}
The encoder is a standard Transformer encoder comprising 6 layers with hidden dimension $h$ equal to 128 and 8 self-attention heads. The encoder receives the context as input (see Fig. \ref{fig:mainfigure}), which consists of a $C\times{3}$ tensor where $C$ is the context length equal to $200$ points -- corresponding to $400 s$ -- and and the second dimensions includes voltage, current, and time. The first two dimensions (voltage and current) of the input matrix are passed to a linear layer, which enhances its dimension from 2 to $h$, resulting in a $C\times{h}$-dimensional tensor $x_{vc}$. The time dimension is fed into a standard positional embedding with dimension $h$, again resulting in a $C\times{h}$-dimensional tensor $x_{t}$. The input $x$ to the first self-attention layer consists in the element-wise sum of $x_{vc}$ and $x_{t}$. The output of the encoder consists then in a $C\times{h}$ tensor. As shown in our experiments, this tensor contains important information about the degradation level of the battery and will be used to condition the decoder in its predictions.
\paragraph{Decoder.}
The decoder is a Transformer decoder comprising 6 layers with hidden dimension $h$ equal to 128 and 8 self-attention heads. However, since the attention mechanism is highly computationally demanding when the length of the input sequence is large, we opportunely modify the input of the decoder to lower the computational requirements of the model. In particular, the input to the decoder is given by the entire current profile for which we wish to predict the corresponding voltage discharge curve. This time series can reach several hours in length, corresponding to thousands of time steps. More precisely, the input consists of an $L\times{1}$ tensor, where $L$ is the length of the current load profile. Since $L$ is usually large, we split it into $T=\frac{L}{n}$ sub-time series, each of length $n = 64$. Each sub-time series is then treated as a separate token of dimension  $n$. To make the tokens match the hidden dimension of the transformer, we pass them through a linear layer which projects them to an $h$-dimensional space. We also use an $h$-dimensional positional embedding, which we sum element-wise with the sequence of tokens. Besides processing the input sequence via self-attention, the decoder additionally performs cross-attention on the output extracted by the encoder. The output of the last self-attention layer will consist of a sequence of $T$ $h$-dimensional tokens. After being linearly projected back to an $n$-dimensional space, this sequence is reshaped into a $T\times{n}$ ($=L$) one-dimensional time series, which represents the final prediction of the network.

\paragraph{A note on Uncertainty Estimation.}
As an additional contribution, we enable our Dynaformer model to perform uncertainty quantification. This is made possible by a very simple procedure: following \cite{gal2016dropout}, we activate the dropout layers present in the encoder and decoder at test time, resulting in a probabilistic model whose output changes at each forward pass. In this way, by performing several forward steps through the model, we obtain a distribution over its predictions. This allows us to perform uncertainty quantification by simply extracting the standard deviation of the so-obtained distribution.

\subsubsection*{LSTM}
The LSTM model is a standard encoder-decoder architecture in which both the encoder and the decoder consist of LSTM cells. As for the Transformer encoder, the input to the encoder is the $C\times{3}$ context tensor. The third dimension, representing time, is first fed into a positional embedding and then concatenated again with the remaining two dimensions. We take the final hidden state and the final cell states as the output of the encoder and use them to initialize the initial hidden and cell states of the decoder.
The decoder is thus conditioned on the information extracted by the encoder via the processing of the context. The decoder receives as input the current profile and outputs the voltage discharge curve for each time step. The hidden dimension of the LSTM cells in both the encoder and the decoder is equal to 1,000, resulting in a model comprising about 8.1 million parameters.
\subsubsection*{FNN}
The FNN architecture is built with a sequence of fully connected layers, with ReLU non-linearity and a final linear layer. The input to the network is the concatenation of the three tensors: 1) the context tensor, which, in contrast to the previously described models, is flattened to a one-dimensional vector consisting of $C\times{3}$ entries; 2) a query point representing the time step at which we want to predict the output voltage; and 3) a single-entry vector containing the value of the constant current profile. The output of the network is the voltage value corresponding to the input query point. The final architecture consists of 5 hidden layers with 1,000 nodes each.
%\subsection{DeepOnet}
%The DeepOnet architecture has been originally introduced in \cite{Lu_2021} to learn nonlinear operators for identifying differential equations. In our case, the operator we aim to learn is that one mapping our context vector onto the full voltage trajectory.\\
%The DeepOnet architecture comprises two fully-connected sub-networks, namely the branch network and the trunk network. ReLU nonlinearities are used between every hidden layer. The $N\times{3}$ dimensional context vector is first passed through a linear layer which projects it to a $N$ dimensional tensor. This is the input to the branch network which, in turn, is made by two hidden layers with 2000 nodes each and a final 500-dimensional output layer. The input to the trunk sub-network is a two-entry vector comprising the value of the constant current profile and the query point at which we want to predict the output voltage. Such a vector is processed by 2 hidden layers with 2000 nodes each and a final linear layer with 500 neurons. The outputs of the branch and trunk networks share the same hidden-dimension, i.e. 500, and the final output can thus be obtained by calculating their dot product. 

\subsection*{Supplementary Note 4: Experimental Setup}\label{expsetup}
In this section, we provide further details on the training process of each method as well as the metrics employed to assess their performance.
\subsubsection*{Training Details}
\paragraph{Computational Resources.} All the methods were trained on 8 GPU GeForce RTX 2080 with 12 GB of memory each. We used Pytorch Lightning \cite{falcon2019pytorch} to train our models in parallel on multiple GPUs. 
\vspace{-0.5 cm}
\paragraph{Training/Validation splits.} We split our dataset into training and validation sets with relative proportions of 85\% and 15\%, respectively. For all models, we saved the checkpoint at which the validation loss reached its minimum value throughout training. 
\vspace{-0.5 cm}
\paragraph{Loss and Optimizer.}
We use the standard mean squared error (MSE) as loss function and the Adam optimizer \cite{adam} with learning rate equal to $10^{-4}$ for all models. We find empirically that higher values of the learning rate lead to instabilities in all the analyzed models. The batch size is the same for all models and is equal to 64.

\paragraph{Training details for Dynaformer and LSTM.}
During training we randomly cropped/extended the current from 55\% to 155\%\footnote{The extension was made by by prolonging the original current profile with its last value.} so that the network could learn not to rely solely  on the current length to predict the EOD time. The length of the context if fixed at $400 s$. Furthermore, we randomly sample the initial time step from which the context is extracted between $0s$ and $90 s$. This was done to make the models less reliant on the initial points of the discharge curve, which can change drastically between different curves. We train the models until validation loss convergence (no improvement for more than 500 epochs). Training time varied significantly between our model and the LSTM, the first converging in about 5 hours, while the latter required more than 1 day.

\paragraph{Training details for FNN.} The FNN was trained until validation loss convergence (no improvement for more than 500 epochs). We fixed the initial time step from which the context is extracted at $90 s$ since otherwise we noticed that training became too unstable.  The training time necessary to reach convergence was about 12 hours.

%While we notice a lot of variance in the validation performance test set for the FNN, and DeepOnet. The transformer and the LSTM were much more stable, especially when trained with constant current profiles.
%We use the standard MSE loss with Adam with a learning rate at 0.0001 for all the training of the models. The batch size was of 64 elements. For the transformer and S2S no normalization was done, while in the case of the FNN and DeepOnet we normalize time by diving it by 1000.
%For the transfomer and the S2S model during training we randomly sampled a trajectory from our training set and randomly cropped/extend the current from 55\% and 155\%, so that the network could learn that the current's length does not contain any information.
%The FNN and DeepOnet were trained until training loss convergence (no improvement for more than 500 epochs). Additionally, FNN and DeepOnet were trained the input x init window fixed to 48, since we found it impossible to train it with additional degree of freedom of the time.
%We used 

\subsubsection*{Metrics}
As stated in the main body, we use two main metrics to evaluate our method, the root-mean-squared error (RMSE) and the relative temporal error (RTE). The first is defined as:
\begin{equation}
    RMSE = \sqrt{\frac{1}{L}\sum_{i=0}^T(y_i - \hat{y}_i)^2}
\end{equation}
where $y_i$ and $\hat{y}_i$ are the ground truth and prediction, respectively, and $L$ is the length of the sequence.\\
As an additional metric, we introduce the RTE. The goal of this metric is to probe the reliance of the algorithms on the length of the input current. Given a fully discharged voltage curve $v$ and the corresponding constant current trajectory $i$, we would expect a model receiving a current $\bar{i}$ longer than $i$ to output a voltage $\hat{v}$ with the same EOD time as $v$ and not a higher one. Conversely, given as input a shorter current, the algorithm has to be able to output a not-discharged voltage curve. We implement the RTE metric by gradually increasing the length of the current profile starting from the 70\% of its original length to 130\%. We prolong the current profile by repeating its last value multiple times. For each length in the above range, we fed the corresponding current into the model and we inspect if and when the predicted voltage curve reaches the end of discharge. We keep the \emph{maximum} error made by the model over all considered time steps. The pseudocode for the RTE calculation is shown in Algorithm. \ref{alg::alg1}.

\begin{algorithm}[H]
\footnotesize
\caption{Relative Temporal Error (RTE) Calculation}\label{alg::alg1}
\begin{algorithmic} 
\REQUIRE Trained model $f$, Dataset $\mathcal{D}=\{i_k,v_k\}_{k=1}^N$, Bounds $\Delta{t_{-}},\Delta{t_{+}}$, Context length $T_c$
\FOR[{Loop over the Dataset}]{$k$ in $\{1..N\}$}
\STATE $t_{0} = len(i_k) = len(v_k)$ 
\STATE $t_{\pm} = t_0 \pm \Delta{t_{\pm}}$
\STATE $E_{(max,\pm)} \leftarrow 0$ 
\FOR[{Loop over the sequence length}]{$t$ in $\{t_{-}...t_{+}\}$}
\IF{$t<t_0$}
\STATE $i_{k,new} = i_k[0,t]$ \hfill\COMMENT{Truncate current length up to $t$}
\STATE $pred = f(i_{k,new},v_k[0,T_c])$
\IF [{Check early EOD predictions}
]{$pred[-1]<3.2$}
\STATE $E_{(max,-)} \leftarrow \max(E_{(max,-)},1- \frac{t}{t_{0}})$ 
\ENDIF
\ELSE
\STATE $i_{add} = repeat(i_k[-1],t-t_0)$
\STATE $i_{k,new} = Conc[i_k,i_{add}]$ \COMMENT{Extend $i_k$ with its last value repeated $t-t_0$ times}
\STATE $pred = f(i_{k,new},v_k[0,T_c])$
\IF {$pred[-1]>3.2{\hspace{0.1cm}}$}
\STATE $E_{(max,+)} \leftarrow (\frac{t}{t_0} - 1)$
\ELSE 
\STATE \textbf{break}
\ENDIF
\ENDIF
\ENDFOR
\STATE $RTE_{k} = \max(E_{(max,-)},E_{(max,+)})$ \COMMENT{Keep the \emph{worst} case error}
\ENDFOR
%\STATE $RTE = \frac{\sum_{k=1}^N(RTE_{k})}{N}$
\RETURN $\{RTE_k\}_{k=1}^N$
\end{algorithmic}
\end{algorithm}

\subsection*{Supplementary Note 5: Additional Results}\label{addresults}
In this section, we present some additional results not reported in the main body. We first focus on the analysis of the performance of the LSTM model on simulated variable current data. Contrary to the LSTM model, we provide further evidence that our approach can deal with long input sequences thanks to the attention mechanism it incorporates. We then perform additional experiments on real data with main intent of investigating whether the output of the Transformer encoder contains meaningful information about the degradation condition of the battery.
%the intent of investigating two main aspects: first we focus on the amount of fine-tuning data needed to improve the performance of the pre-trained model; 2) we analyze the output of the Transformer encoder to investigate whether it contains information about the degradation conditions.

\subsubsection*{Simulated Data}\label{addr}

\paragraph{Variable Current Profiles.}
Supplementary Fig. \ref{fig:lstmshortlong} shows the results for the LSTM baseline on the variable current profiles.  The LSTM baseline is generally outperformed by our method. We investigated the cause, and we observed that the baseline's performance increases drastically when we consider only trajectories with fewer than 4,000 seconds (about 1 h). Models based on recurrent neural networks are known to suffer from vanishing gradients when long time series are used as input and we suspect that this behaviour is a direct manifestation of this phenomenon. On the other hand, our model does not seem to be significantly affected by this issue, as can be seen from Supplementary Fig. \ref{fig:transshortlong}. This is due to the attention mechanism, which can effectively keep track of long-term dependencies in the input sequence without occurring in vanishing gradients.

\subsubsection*{Real Data}
\paragraph{Implicit Ageing Inference.}
In this section, we explore whether our trained encoder contains any information on the degradation level of the input battery data. We repeat the same procedure as in Section \ref{interpret}, that is, we calculate the first principal components of the encoder output. However, this time, since we are testing the algorithm on real data, we do not have access to the explicit degradation parameters associated with each curve. Fortunately, though, we can use the information of the specific cycle each battery discharge curve belongs to. In Supplementary Fig. \ref{fig:phyreal} we plot the two first principal components and we color each point according to the specific cycle of the input data. We repeat this procedure for each of the four real battery datasets.
As shown by the figures, there is a large correlation between the level of degradation of the battery and the position of the embedding. This experiment demonstrates that the encoder is able to extract useful information on the degradation level not only of simulated data, but also of real battery data.

\section*{Supplementary Figures}

\setcounter{figure}{0}  
\captionsetup[figure]{labelformat={default},labelsep=period,name={Supplementary Fig.}}

\begin{figure}[H]
    \centering
    \includegraphics[scale=0.3]{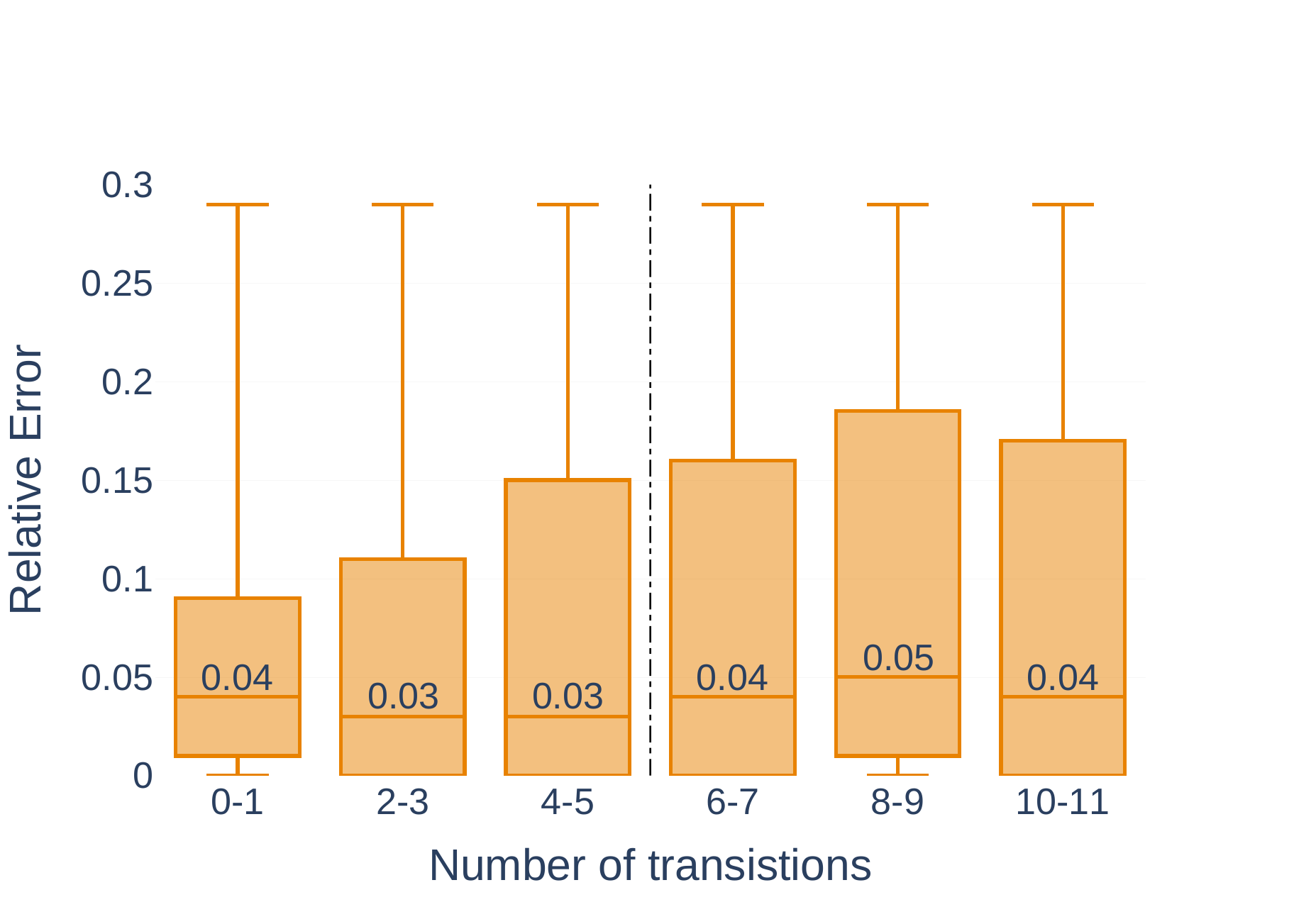}
    \includegraphics[scale=0.3]{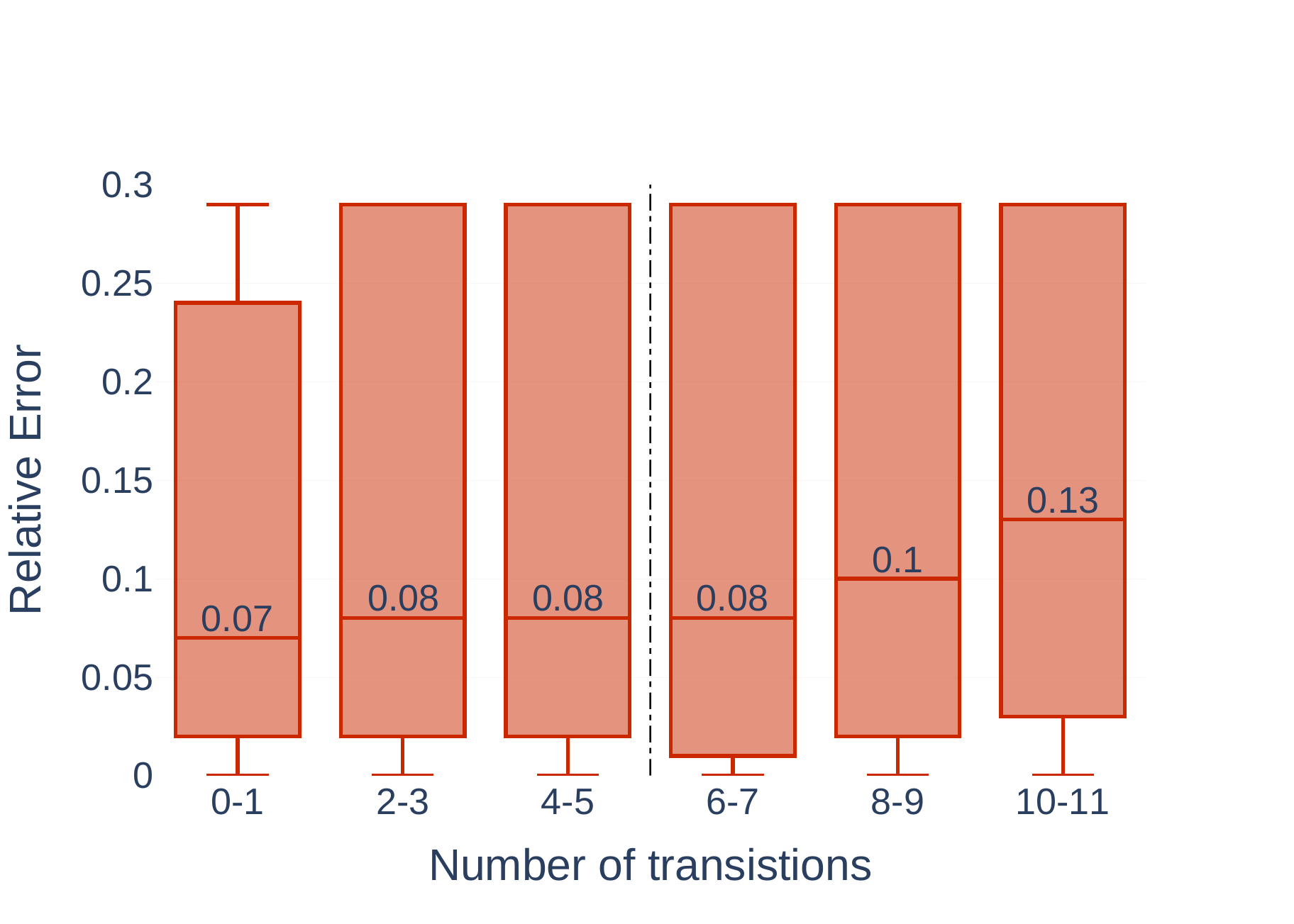}
    \includegraphics[scale=0.3]{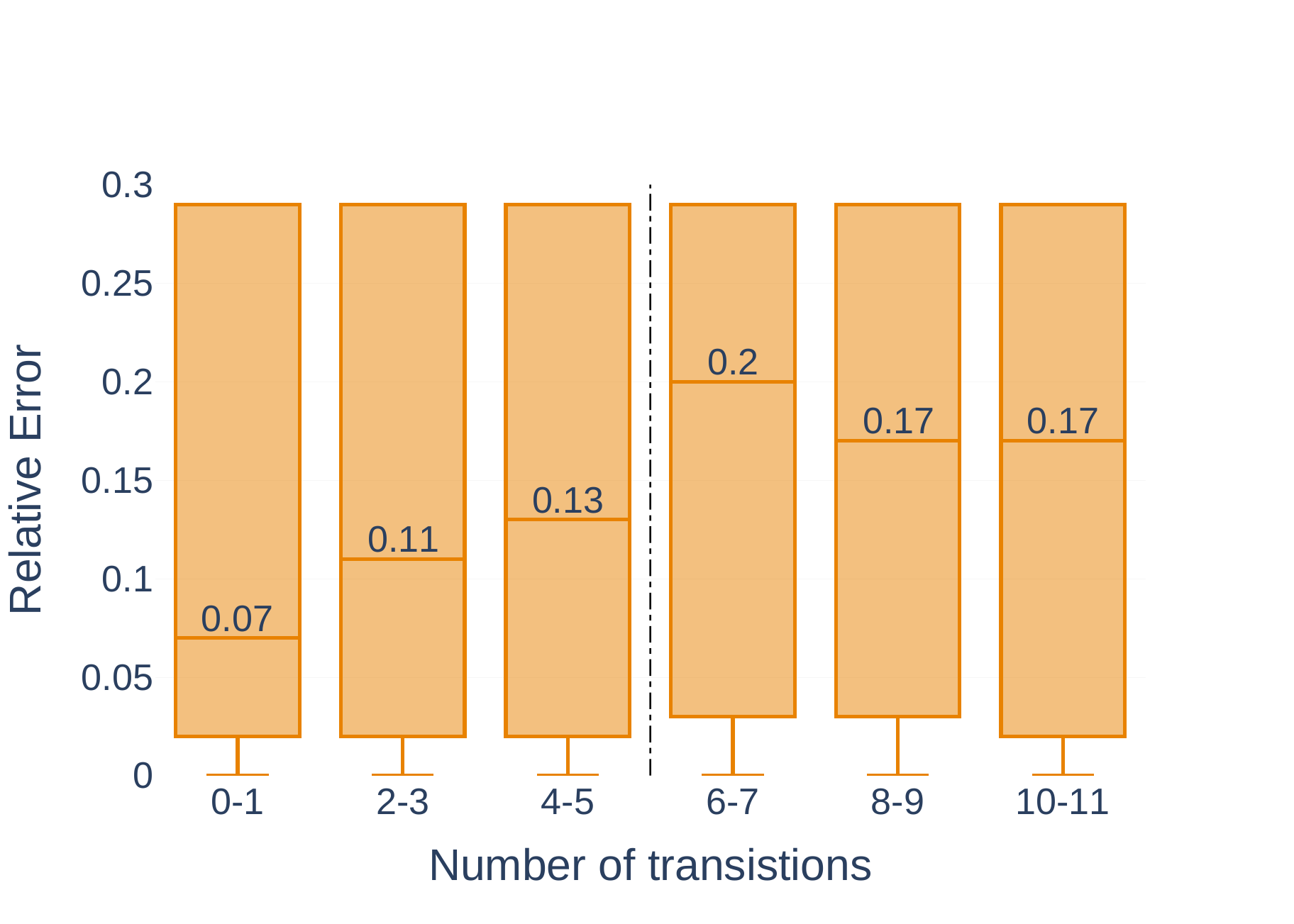}
    \includegraphics[scale=0.3]{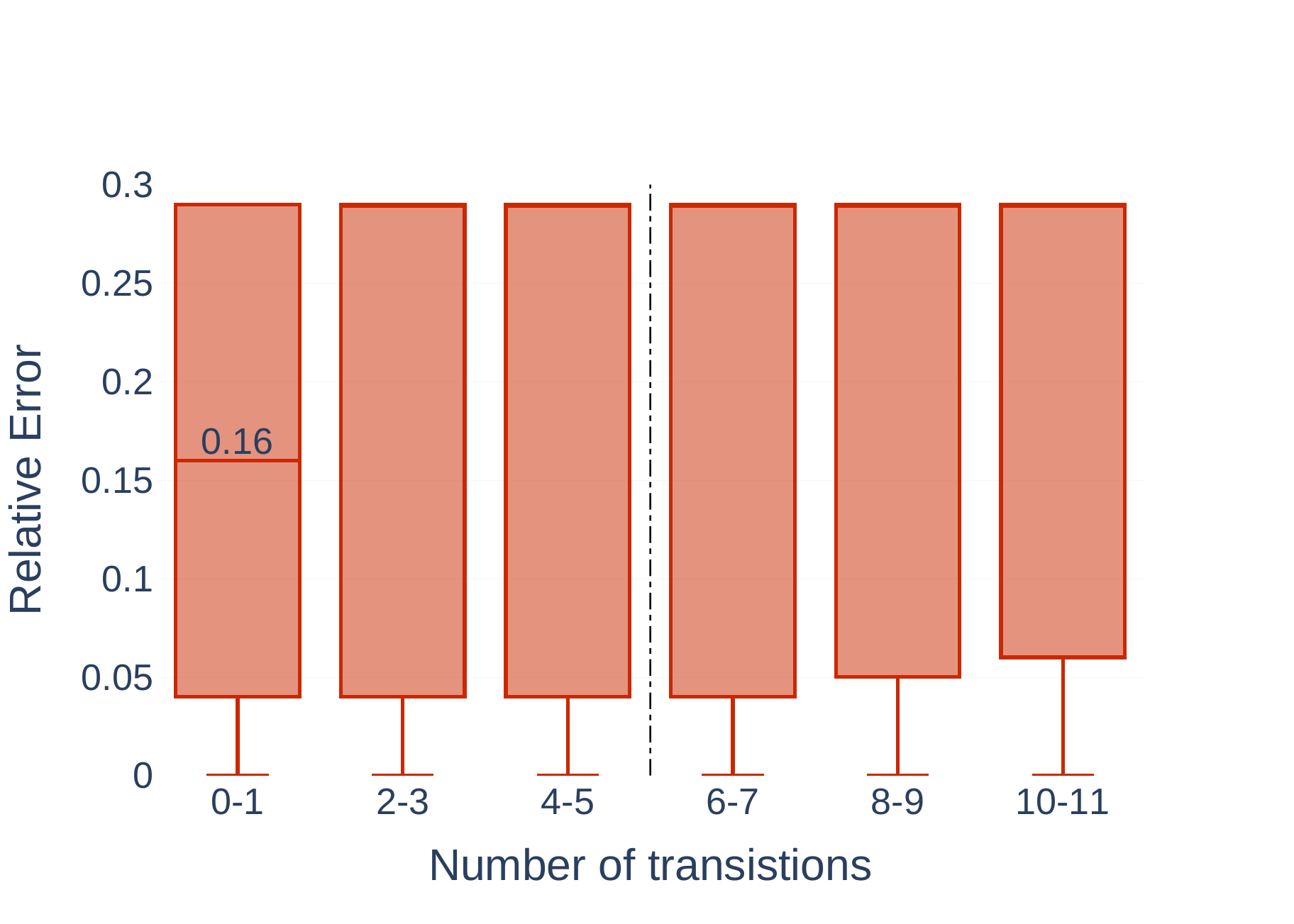}
    \includegraphics[scale=0.3]{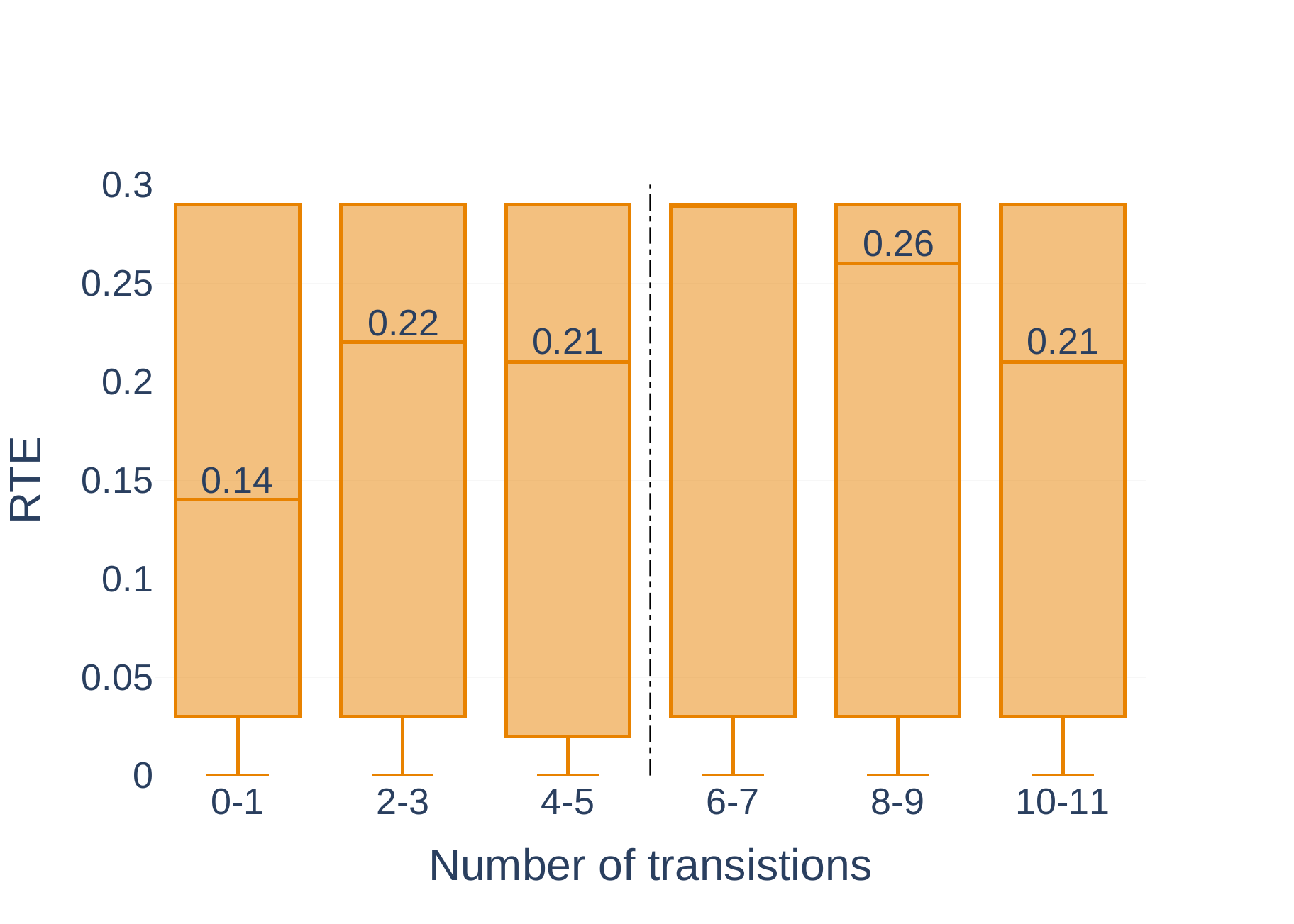}
    \includegraphics[scale=0.3]{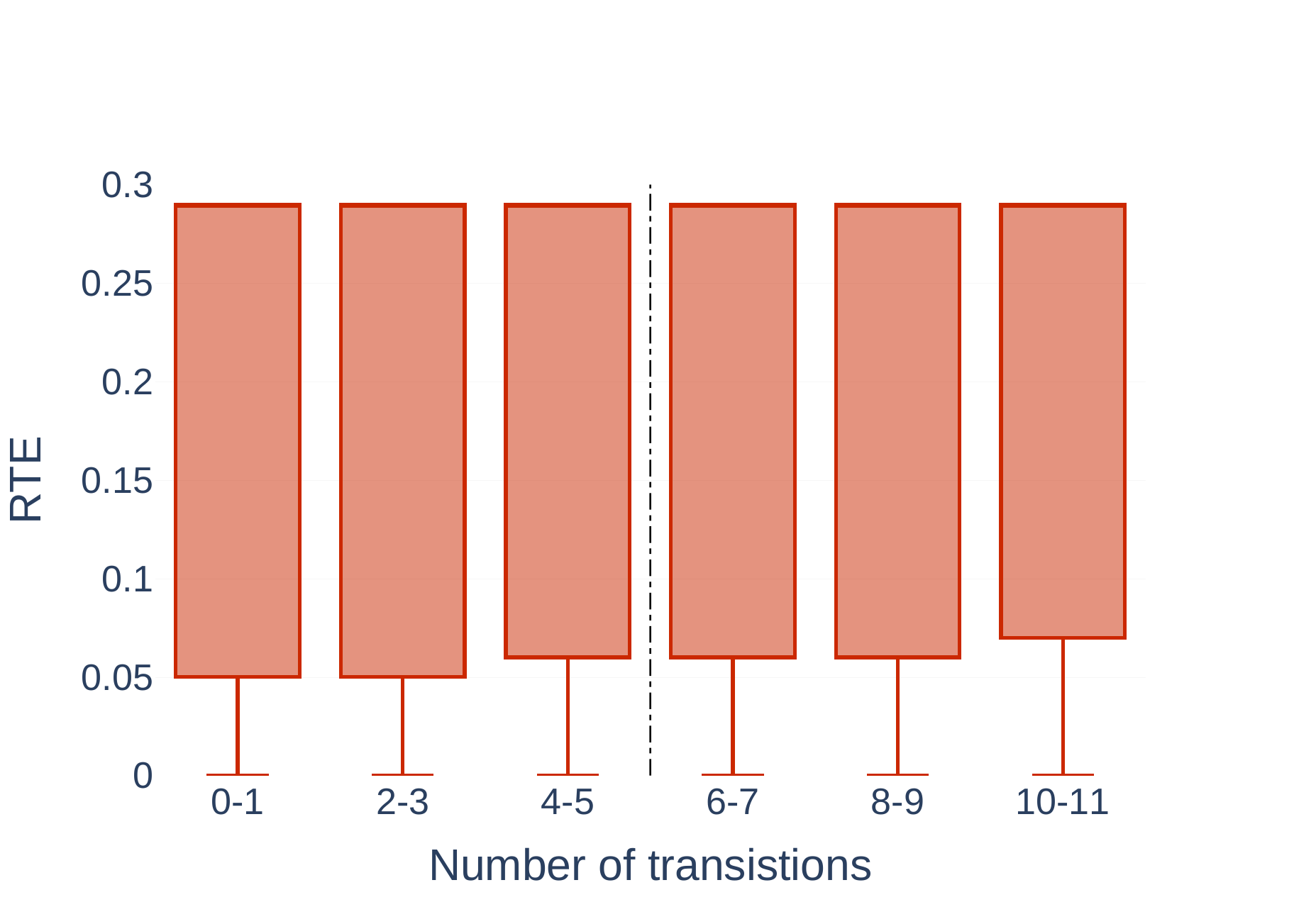}
    \caption{\textbf{LSTM on variable current profiles with different lengths.} (Top) LSTM on current profiles up to 4,000 seconds long.
    (Middle) LSTM on current profiles up to 8,000 seconds long.
    (Bottom) LSTM on current profiles up to 16,000 seconds long. }
    \label{fig:lstmshortlong}
\end{figure}

\begin{figure}[H]
    \centering
    \includegraphics[scale=0.3]{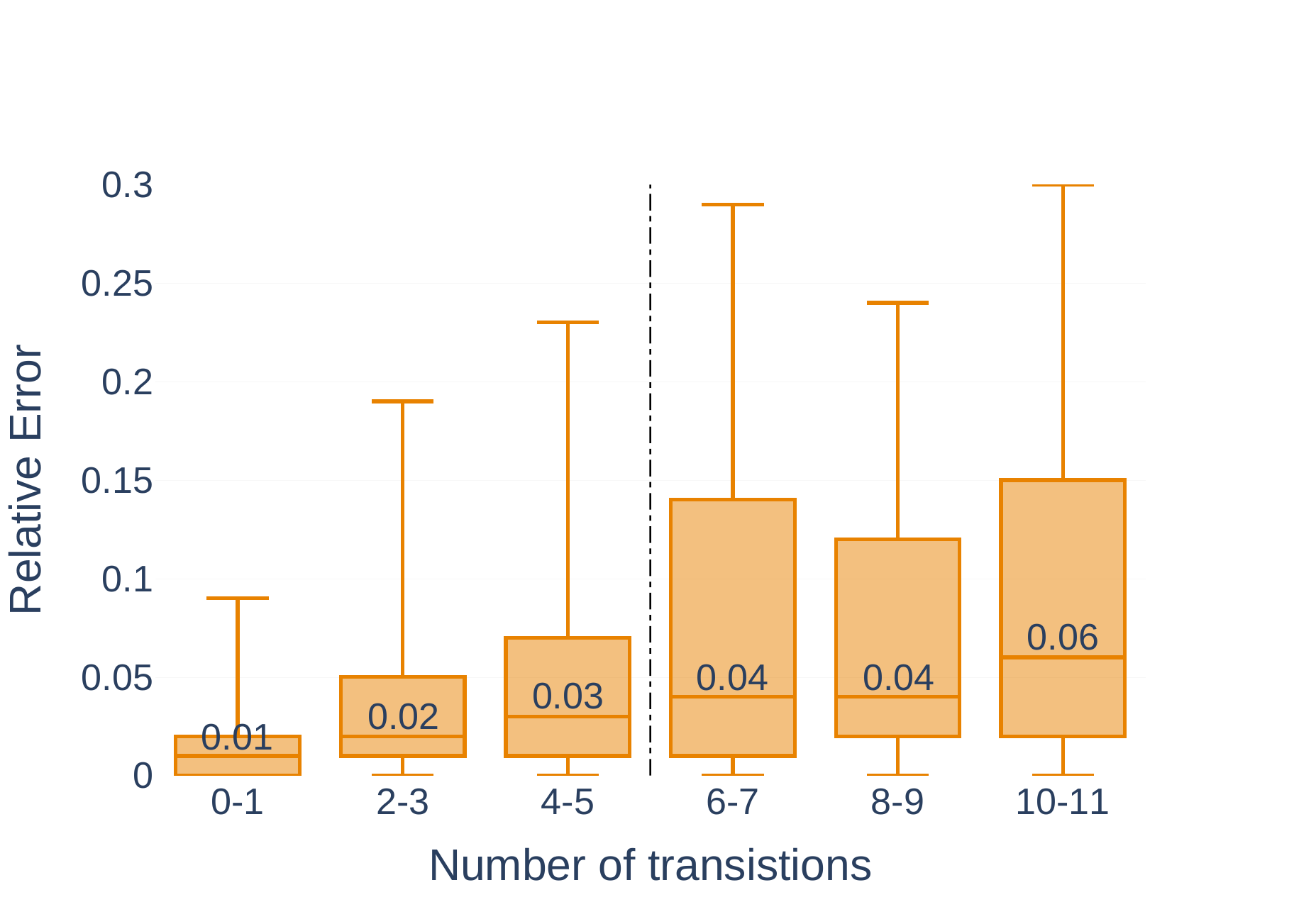}
    \includegraphics[scale=0.3]{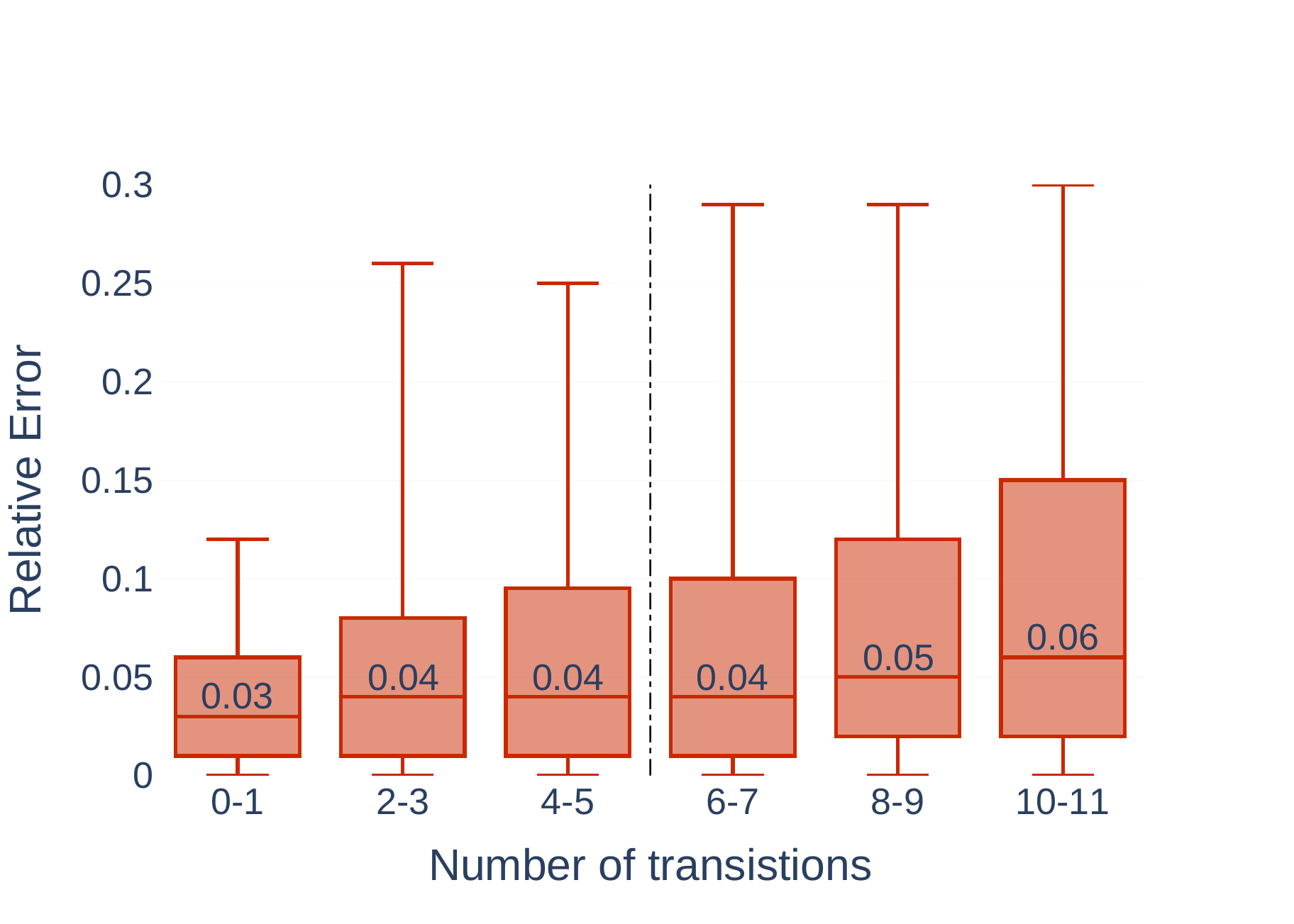}
    \includegraphics[scale=0.3]{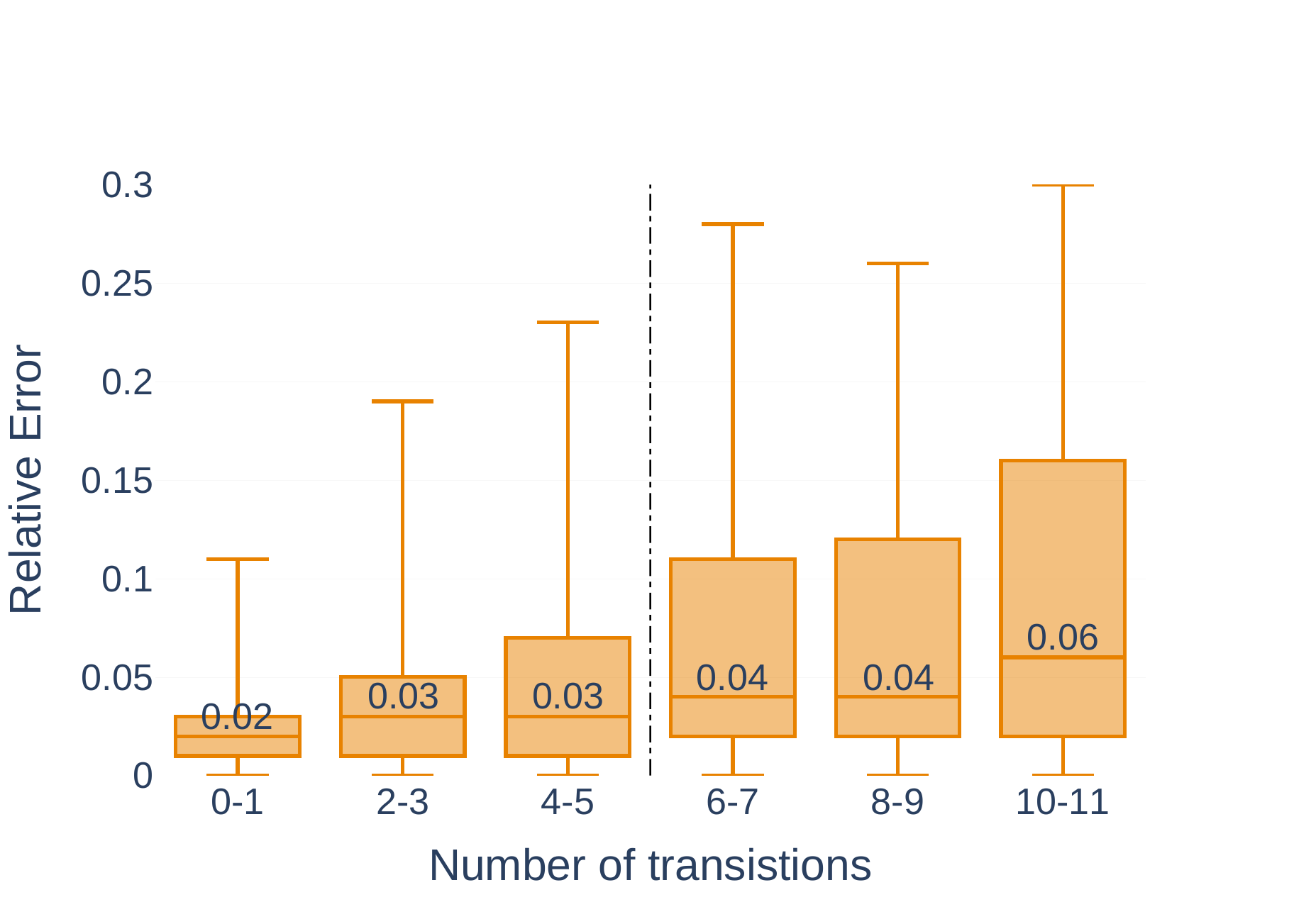}
    \includegraphics[scale=0.3]{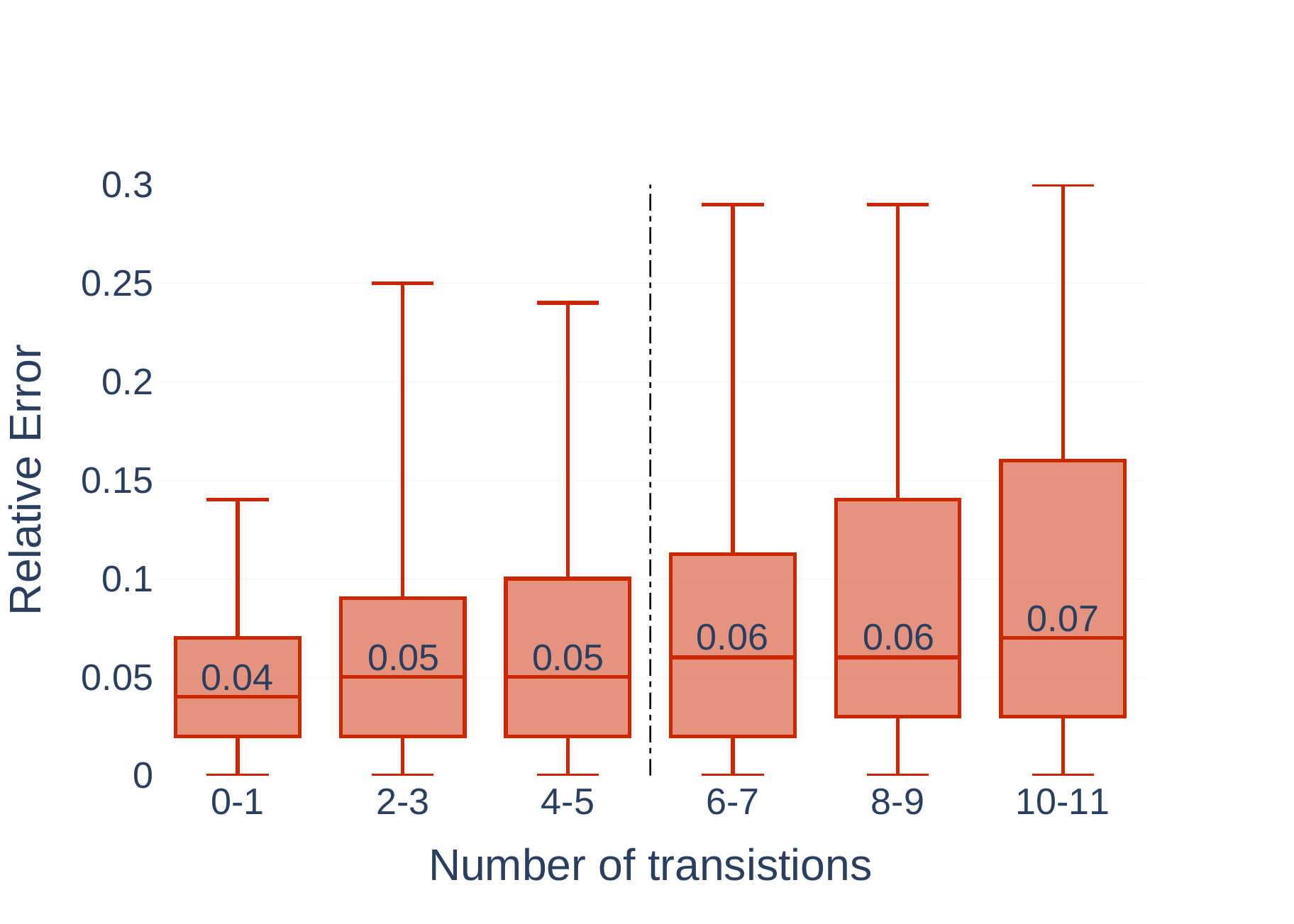}
    \includegraphics[scale=0.3]{results_variable_interpolation_8000.pdf}
    \includegraphics[scale=0.3]{results_variable_extrapolation_8000.pdf}
    \caption{\textbf{Dynaformer on variable current profiles with different lengths.} (Top) Dynaformer on current profiles up to 4,000 seconds long.
    (Middle) Dynaformer on current profiles up to 8,000 seconds long. (Bottom) Dynaformer on current profiles up to 16,000 seconds long (same as Fig. \ref{fig:currents_variable}).}
    \label{fig:transshortlong}
\end{figure}

\begin{figure}[H]
    \centering
    \includegraphics[scale=0.39]{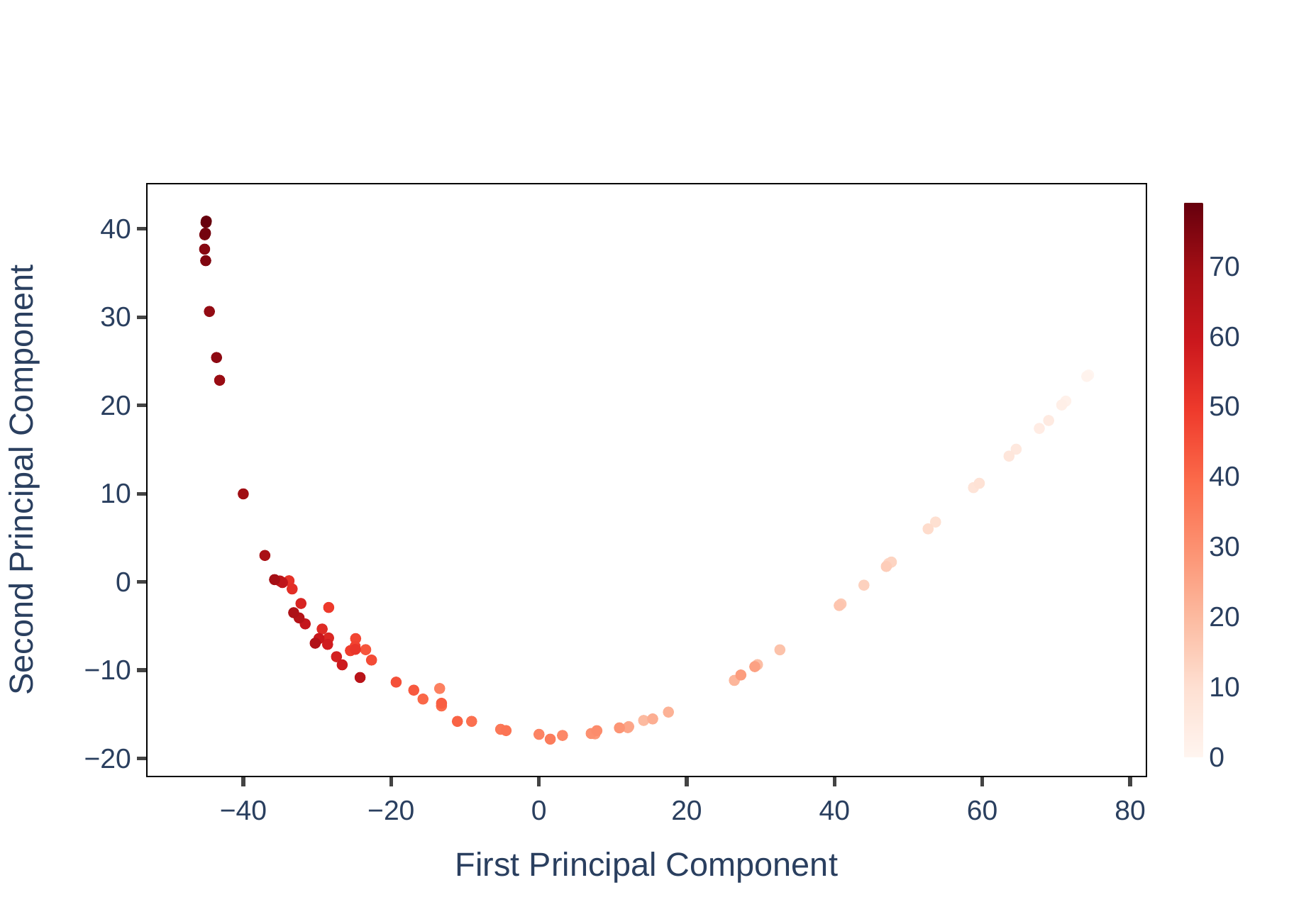}
    \includegraphics[scale=0.39]{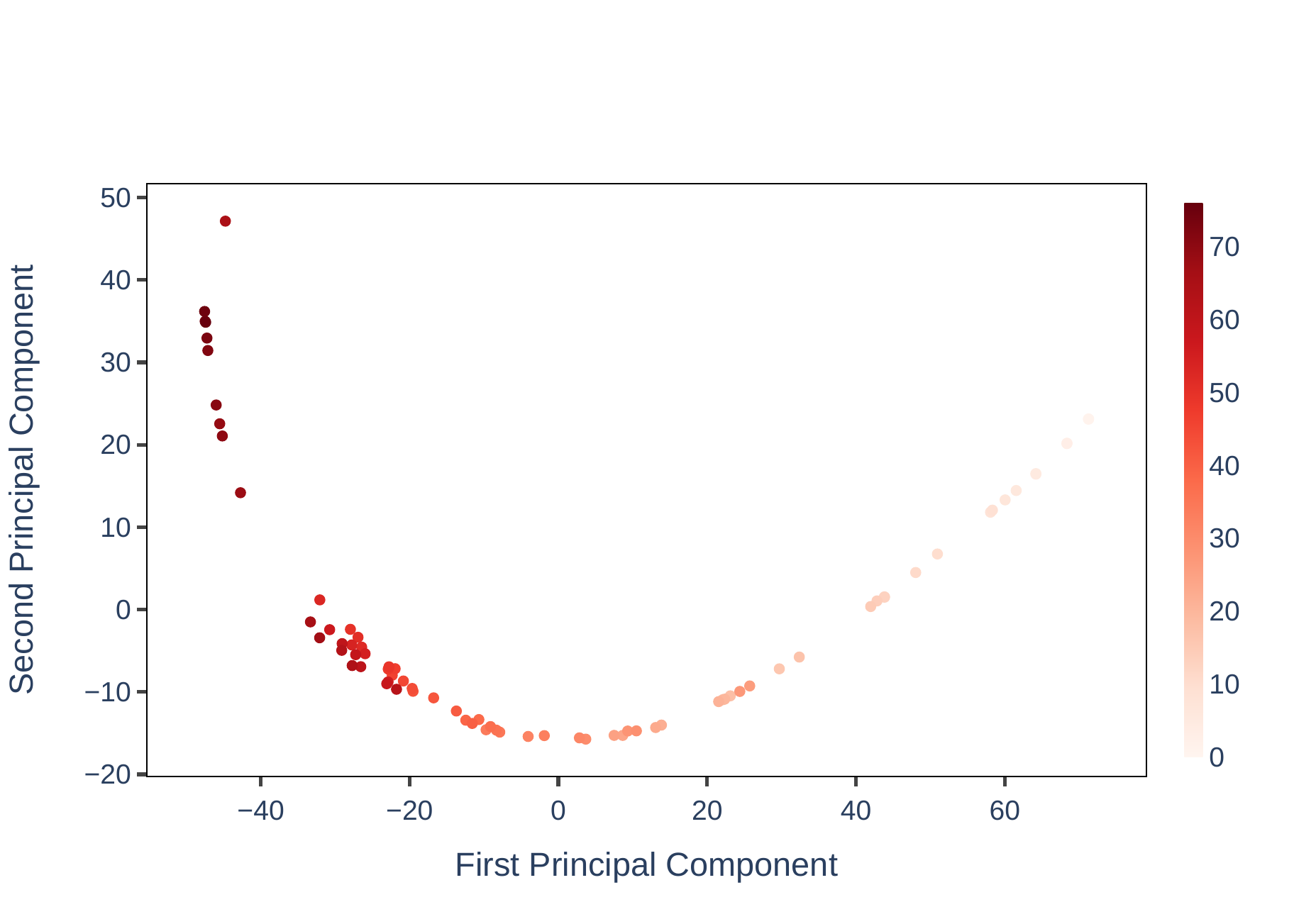}
    \includegraphics[scale=0.39]{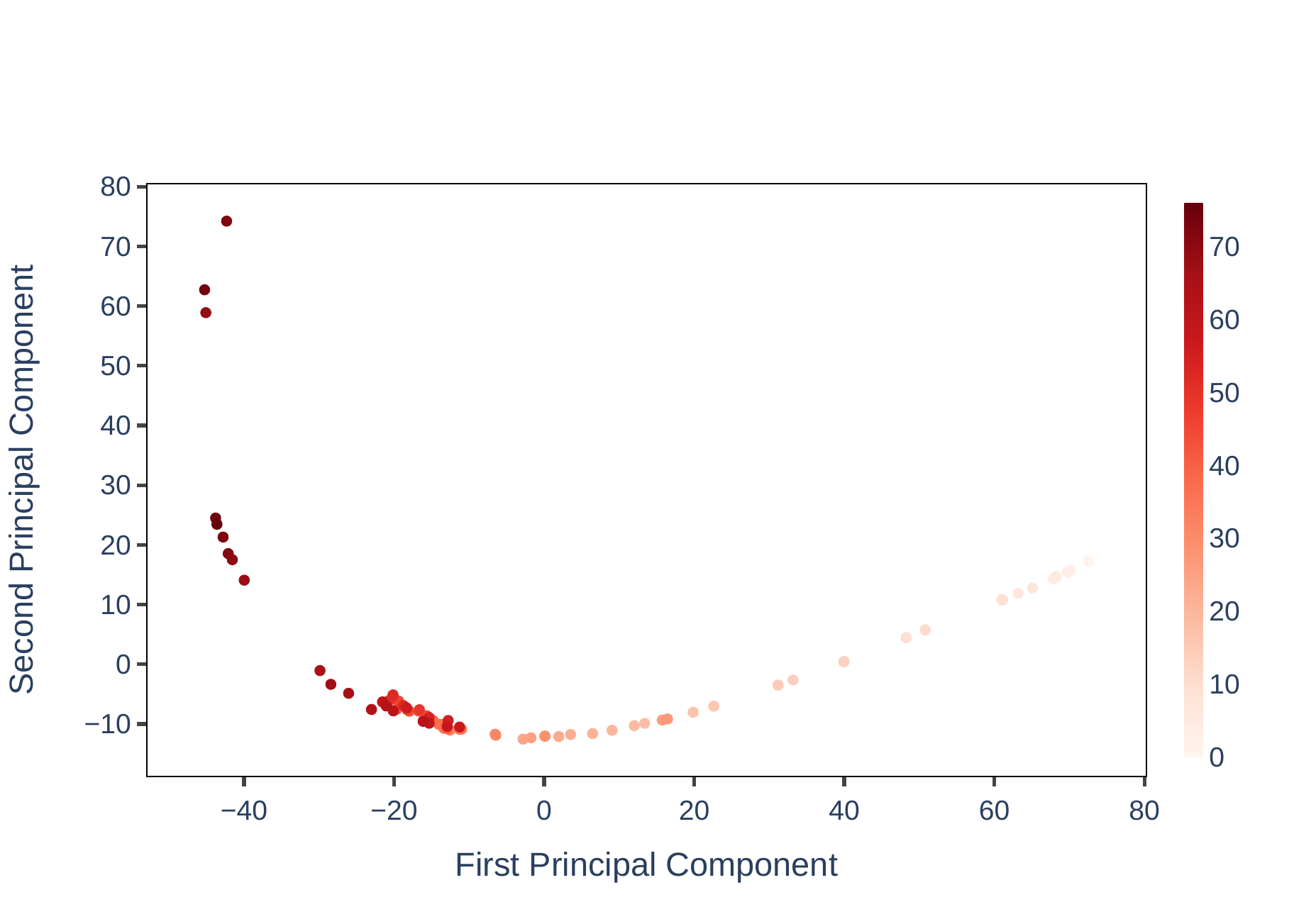}
    \includegraphics[scale = 0.39]{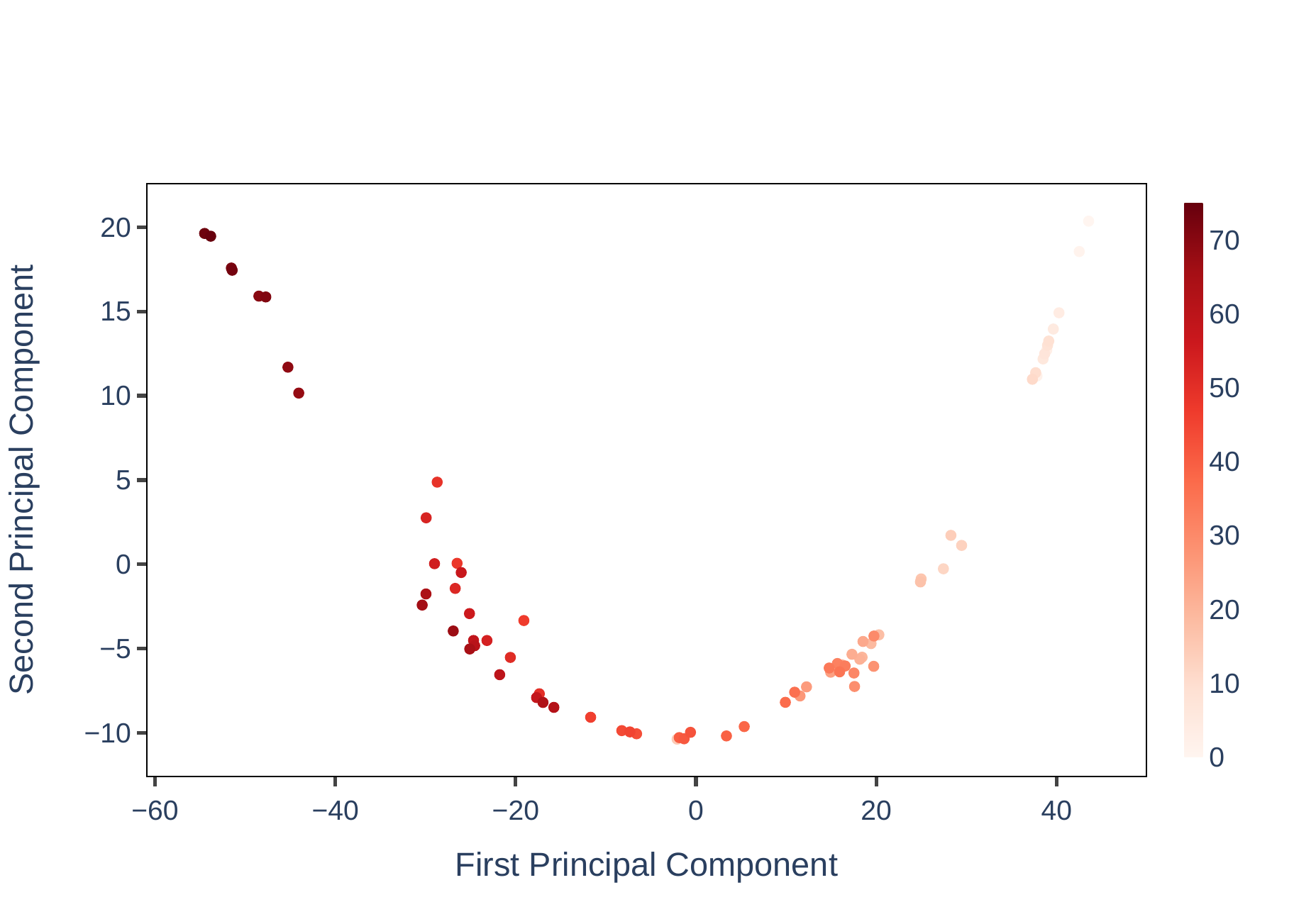}
    \caption{\textbf{Implicit Ageing Modelling on Real Batteries.} (Top) From left to right: results for RW9 and RW10. (Bottom) From left to right: results for RW11 and RW12.}
    \label{fig:phyreal}
\end{figure}
\noindent

\end{document}